\documentclass[lettersize,journal]{IEEEtran}
\usepackage[dvipsnames]{xcolor}
\usepackage{amsmath,amsfonts}
\usepackage{algorithmic}
\usepackage{algorithm}
\usepackage{array}
\usepackage[caption=false,font=normalsize,labelfont=sf,textfont=sf]{subfig}
\usepackage{textcomp}
\usepackage{stfloats}
\usepackage{url}
\usepackage{verbatim}
\usepackage{graphicx}
\usepackage{cite}

\newcommand{\model}{RConE}
\usepackage[hidelinks]{hyperref}
\usepackage{colortbl}
\usepackage{cases}
\newcommand{\submodel}{RConE Engine}
\usepackage{csquotes}
\usepackage{multirow}
\definecolor{Gray}{gray}{0.85}
\newcommand{\mycc}{\cellcolor{Gray}}
\newtheorem{definition}{Definition}
\usepackage{amssymb}
\usepackage{tabularx, booktabs}

\begin{document}
\bstctlcite{bstctl:nodash}

\title{\model: Rough Cone Embedding for Multi-Hop Logical Query Answering on Multi-Modal Knowledge Graphs}
\author{Mayank Kharbanda, Rajiv Ratn Shah, Raghava Mutharaju
\thanks{M. Kharbanda, R. Shah, and R. Mutharaju are with IIIT-Delhi, India.}}

\maketitle

\begin{abstract}
  Multi-hop query answering over a Knowledge Graph (KG) involves traversing one or more hops from the start node to answer a query. Path-based and logic-based methods are state-of-the-art for multi-hop question answering. The former is used in link prediction tasks. The latter is for answering complex logical queries. The logical multi-hop querying technique embeds the KG and queries in the same embedding space. The existing work incorporates First Order Logic (FOL) operators, such as conjunction ($\wedge$), disjunction ($\vee$), and negation ($\neg$), in queries. Though current models have most of the building blocks to execute the FOL queries, they cannot use the dense information of multi-modal entities in the case of Multi-Modal Knowledge Graphs (MMKGs). We propose \model{}, an embedding method to capture the multi-modal information needed to answer a query. The model first shortlists candidate (multi-modal) entities containing the answer. It then finds the solution (sub-entities) within those entities. Several existing works tackle path-based question-answering in MMKGs. However, to our knowledge, we are the first to introduce logical constructs in querying MMKGs and to answer queries that involve sub-entities of multi-modal entities as the answer. Extensive evaluation of four publicly available MMKGs indicates that \model{} outperforms the current state-of-the-art. The source code and datasets are available at \url{https://github.com/kracr/rcone-qa-mmkg}.
\end{abstract}


\section{Introduction}\label{sec:introduction}
\IEEEPARstart{A}{} Knowledge Graph (KG)~\cite{hogan2021knowledge} is a directed graph with a set of entities (nodes) and directed relations among those entities (edges). KGs are an excellent tool for representing data in graph topology. They are used in applications such as question-answering and recommendation systems in diverse fields like biomedicine, physics, and geoscience~\cite{ji2021survey}.

Multi-hop logical query answering over KGs has gained attention recently~\cite{zhang2021neural}. Various neural methods are proposed to answer a logical query. It involves traversing one or more hops over a KG to reach the answers. The query generally consists of First Order Logic (FOL) operators, such as existential quantification ($\exists$), conjunction ($\wedge$), disjunction ($\vee$), and negation ($\neg$).

\textbf{Current State-of-the-Art (SOTA).} There are two major challenges to consider while handling logical query answering over KGs~\cite{ren2020query2box}. First, long and complex queries on large KGs incur exponential computational costs. Second, a robust model for handling missing edges in the graphs. Several methods have been proposed to embed the KG and queries in the same space to tackle these issues~\cite{ren2023neural}. These methods iteratively progressed to incorporate different FOL operators in the queries. Geometric~\cite{ren2020query2box,zhang2021cone}, and probabilistic~\cite{ren2020beta} methods embed the queries as geometrical shapes and probabilistic distributions, respectively. These methods are scalable and do not keep track of the intermediate nodes.

\textbf{Shortcomings of SOTA Methods.} \textbf{[RQ1]} Multi-Modal Knowledge Graphs (MMKGs) are KGs with multiple modalities, such as images, texts, and videos, as entities \cite{zhu2022multi}. Though the current approaches can handle all the FOL operators well, they cannot incorporate the rich information of multi-modal entities. The node's embedding, in these models, is based on the relations with its neighbors. They do not consider the entity’s features, which may lead to the loss of critical information in the case of multi-modal entities. \textbf{[RQ2]} There can be multiple subjects in a single multi-modal entity, and it might be that all the subjects are not answers to a query. One of the goals of this work is to get those individual subjects as answers. \textbf{[RQ3]} One way to tackle [RQ1] and [RQ2] is to generate a sub-KG for each multi-modal entity before training (offline). Convert the original MMKG to a non-multi-modal KG by merging all sub-KGs with the MMKG. However, for large MMKGs, constructing the sub-KGs for each multi-modal entity will incorporate high pre-processing and space overhead.

To address these challenges, we propose \model, an embedding method for answering logical queries on MMKGs. In this work, we focus on entity/relational labels and images as the modalities\footnote{From here on, we refer to image entities as multi-modal entities.}. We summarize our contributions as follows.

\begin{figure}[!t]
    \centering
    \includegraphics[width=\columnwidth]{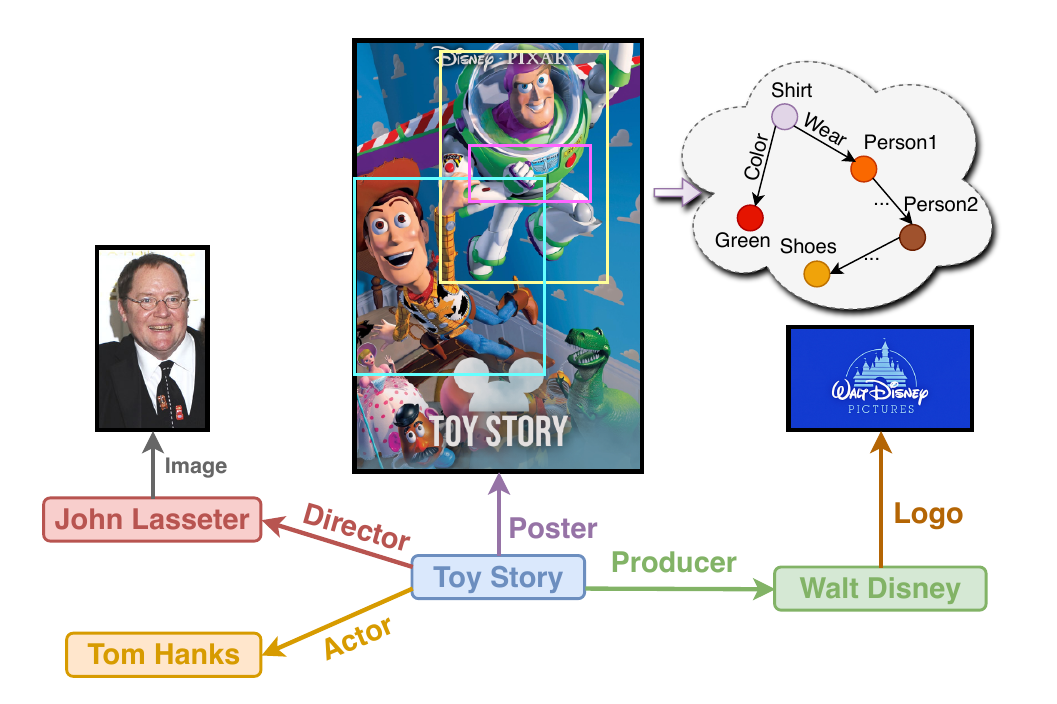}
    \caption{A toy example of a Multi-Modal Knowledge Graph, consisting of images (\emph{Toy Story (Poster)}, \emph{John Lasseter (Image)}, and \emph{Walt Disney (Logo)}) as a modality, along with regular entities. The cloud consists of scene graph generated from entities (rectangular boxes) in the \emph{Toy Story (Poster)}.}
    \label{fig:mmkg}
\end{figure}

\begin{figure}[!t]
    \centering
    \includegraphics[width=0.52\textwidth]{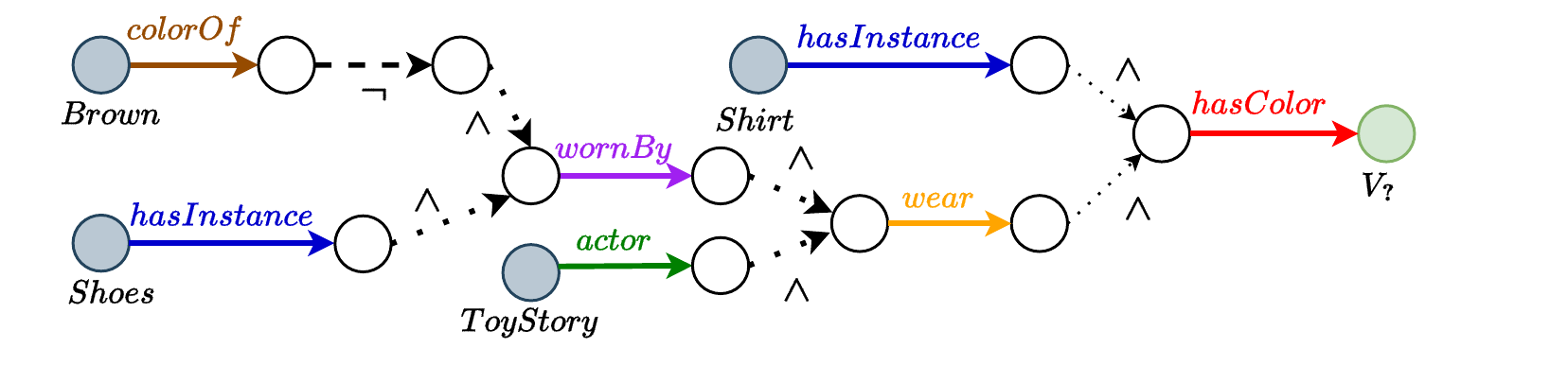}
    \caption{Computational graph for the FOL query $q=V_{\boldsymbol{?}}.\boldsymbol{\exists} V_1, V_2, V_3: ((({\neg\color{brown}colorOf}  (Brown, V_1) \boldsymbol{\wedge} {\color{blue}hasInstance}(Shoes, V_1) )\boldsymbol{\wedge}{\color{Purple}wornBy}(V_1, V_2) \boldsymbol{\wedge} {\color{ForestGreen}actor}(ToyStory, {V_2})) \boldsymbol{\wedge} {\color{orange}wear}(V_2, V_3) \boldsymbol{\wedge}{\color{blue}hasInstance}(Shirt, V_3))\boldsymbol{\wedge}{\color{red}hasColor}(V_3, V_{\boldsymbol{?}})$, corresponding to the question \enquote{Shirt color of the actor not wearing brown shoes in the Toy Story movie.} Here nodes represent the entity set, and edges are the logical operations. The green node ($V_?$) is the answer to the query.}
    \label{fig:compute_graph}
\end{figure}

\textbf{Contribution I: Logical Query Answering on MMKGs at a finer granularity.} We propose a novel problem of query answering using FOL constructs on MMKGs. In this problem, the answers might not be complete entities but some part (sub-entities) of the multi-modal entities. To our knowledge, we are the first to handle such queries. Consider the MMKG in Figure \ref{fig:mmkg}. For the query, \enquote{Shirt color of the actor not wearing brown shoes in the Toy Story movie,} the answer is \emph{Green}. The related computational graph is in Figure \ref{fig:compute_graph}. In the MMKG, \emph{Green} (shirt), is in the \emph{Toy Story (Poster)} entity (sub-entity of \emph{Toy Story (Poster)} entity), but there is no entity \emph{Green} in the MMKG. To cater to these kinds of queries, we propose our model \model.

\textbf{Contribution II: \model{} – A novel perspective of handling the query answering on MMKGs.} We extend one of the neural query answering methods, ConE \cite{zhang2021cone}, to address the multi-modal entities. The proposed method helps embed entities and sub-entities of an MMKG in the same space where we embed the query.

\textbf{Contribution III: MMKG Query Generation.}
We extend a logical query generation algorithm~\cite{ren2020beta} to incorporate the multi-modal entities/sub-entities while generating the training queries. We create queries based on 14 structures described in the literature.

\textbf{Contribution IV: Extensive Evaluation.} We extensively test our method over four multi-modal datasets. The difference between \model{} and the best-performing baseline is $56.5\%$ in the MRR score for one of the datasets.

\section{Related Work}\label{sec:Literature}
{\setlength{\parindent}{0cm}\textbf{Multi-Hop Query Answering.}
There are two types of multi-hop querying techniques: path-based answering \cite{xiong2017deeppath,lin2023multi,chen2019embedding,guo2018knowledge,guo2016jointly,wang2020entity} and logical query answering. In path-based answering, rules or paths of the KG are traversed to answer a query. It is generally used to improve the link prediction task. Logical query answering~\cite{ren2023neural} embeds the KG and the query in the same space to evaluate the answer set. Our proposed model is based on logical query answering.}

GQE~\cite{hamilton2018embedding} proposed the initial query engine. The model answers the query with the closest answer embedded to the query. Query2Box~\cite{ren2020query2box} extended the work by representing the region of interest (ROI) using d-dimensional boxes. This ROI helps incorporate multiple answers to a query. The model could handle the disjunction ($\vee$) and conjunction ($\wedge$) operators. BetaE~\cite{ren2020beta} and ConE~\cite{zhang2021cone} extended Query2Box to integrate the negation ($\neg$) operator, making them fully compatible with FOL operators. While BetaE uses $d$-beta distributions to represent a query component, ConE uses $d$-ary sector cones. CylE~\cite{nguyen2023cyle} embeds the query in three-dimensional cylinders rather than 2d sector-cones. GNN-QE~\cite{zhu2022neural} uses NBF-Net~\cite{zhu2021neural} as a link prediction algorithm for projection and fuzzy logic for FOL operators. Extensions have also been proposed for hyper-relational KGs~\cite{alivanistos2021query} and temporal KGs~\cite{lin2022tflex}. A detailed survey on different query-answering methods is provided in~\cite{liang2022survey}. Information about the current state of the art and the prospective future of logical query answering over KGs is contemplated in~\cite{ren2023neural}.

{\setlength{\parindent}{0cm}\textbf{Question-Answering on MMKGs.}
There are several works on path-based question-answering over an MMKG. Single-hop models like IKRL~\cite{xie2016image} use attention to capture the visual features of images. TransAE~\cite{wang2019multimodal} adds multi-modal features and extends the conventional single-hop embedding model TransE~\cite{bordes2013translating}. MMKGR~\cite{zheng2022mmkgr} proposed the initial multi-hop path-based query answering method using reinforcement learning for a unified gate attention network to filter noise. A review of the application of MMKGs in different fields is presented in~\cite{zhu2022multi}. Though these works use MMKGs to evaluate a query, they do not support logical queries with FOL constructs.} 

\section{Preliminaries}\label{sec:prelim}

{\setlength{\parindent}{0pt}\textbf{MMKG.}
A Multi-Modal Knowledge Graph $G(\mathcal{V},\mathcal{R},\mathcal{U}, \mathcal{M})$ is a directed graph where $\mathcal{V}$ and $\mathcal{R}$ are entity and relation sets, respectively. $\mathcal{U}$ is the triplet set, represented as}
\begin{equation}
    \mathcal{U}=\{(e_s,r,e_d)\;\;|\;\;e_s,e_d \in \mathcal{V}, r \in \mathcal{R}\}
\end{equation}
There are $|\mathcal{M}| = k$ modalities in the graph, such that 
\begin{equation}
    \gamma(e_i) \in \mathcal{M} = \{m_1, m_2, \ldots, m_k\}, e_i \in \mathcal{V}
\end{equation}
where $m_1$ is the generic entity label. In Figure~\ref{fig:mmkg}, the entities 
$\{$Tom Hanks, Walt Disney$\} \in m_1$ (generic) and $\{$Toy Story (Poster), Walt Disney (Logo)$\} \in m_2$ (images). We consider $k=2$, in this work.

{\setlength{\parindent}{0pt}\textbf{Sub-Entity Knowledge Graph (Scene Graph).} For each node, $e_j$, such that $\gamma(e_j) = m_2$, we define a sub-entity KG or a scene graph as $SG_j(\mathcal{V}_j, \mathcal{R}_j, \mathcal{U}_j)$. $\mathcal{V}_{j}$, $\mathcal{R}_j$, and $\mathcal{U}_j$ are sub-entity, sub-relation, and sub-triplet sets, respectively. The $SGs$ are the KGs describing each multi-modal entity in $G$.}

{\setlength{\parindent}{0pt}\textbf{FOL Queries.} First Order Logic (FOL) queries $q \in Q = \{Q_{train}, Q_{test}\}$ consists of logical operators such as conjunction ($\wedge$), disjunction ($\vee$), existential quantification ($\exists$), and negation ($\neg$). We are not including universal quantification ($\forall$), similar to~\cite{ren2020beta, zhang2021cone}, as its applications in real-world KGs are rare.  Queries are expected to be in Disjunctive Normal Form (DNF). This enables us to handle the union operator at the end, which helps the model to be scalable for long queries. An FOL query $q \in Q$, with respect to an MMKG, comprises a non-variable anchor entity set
\begin{equation}
    \mathcal{V}_a\subseteq \mathcal{V_{\cup}} = \Biggl\{\bigcup_{j=1}^{|\mathcal{V}|} \mathcal{V}_j \cup \mathcal{V} \bigm| \gamma(e_j) = m_2\Biggr\}
\end{equation}
along with existentially quantified bound variables $V_1, \ldots, V_k$ and a target variable $V_?$ (query's answer). The DNF for the query is
\begin{equation}
    q[V_?]=V_?.\exists V_1, \dots, V_k : c_1 \vee c_2 \vee ... \vee c_n
\end{equation}
with $c_i=\{l_{i1}\wedge l_{i2} \wedge ... \wedge l_{il}\}$, being conjunctions of one or more literals $l$, such that $l_{ij} = r(v_a,V)$ or $\neg r(v_a,V)$ or $r(V',V)$ or $\neg
r(V',V)$, where $v_a \in \mathcal{V}_a$, $V \in \{V_?,V_1, \ldots ,V_k\}$, $V' \in
\{V_1,...,V_k\}$, $V \neq V'$, and $r \in \mathcal{R}_\cup = \{\bigcup_{j=1}^{|\mathcal{V}|} \mathcal{R}_j \cup \mathcal{R} \;|\; \gamma(e_j) = m_2\}$.}

The objective of a FOL query answering model for an MMKG is
\begin{equation}
    \theta' = \arg\min_{\theta}\sum_{q \in Q_{test}}  \mathcal{L}(g_{\theta}(G, q), g_{\theta}(\mathcal{A}_{gt}(q)))
\end{equation}
Here, $\theta$ is the set of parameters for the embedding module $g$, $\mathcal{A}_{gt}(q)$ is the ground-truth answer set of query $q$, and $\mathcal{L}$ is the loss function.

{\setlength{\parindent}{0cm}\textbf{Query Answer Types.} Through \model{}, we will handle two sets of answers to a query. \textbf{Type (I) answers}: For query Q1 -- \enquote{Shirt color of the actor not wearing brown shoes in the Toy Story movie,} while there is no entity for the color \emph{Green} in $\mathcal{V}$ (Figure \ref{fig:mmkg}), it can be extracted from the multi-modal entity, the \emph{Toy Story (Poster)}. Hence, it is a sub-entity (not an entity) answer in MMKG. \textbf{Type (II) answers}: For query Q2 -- \enquote{Director of the movie Toy Story.} we have an entity, \emph{John Lasseter}, in $G$. Hence, the complete entity is the answer to the query.}

{\setlength{\parindent}{0pt}\textbf{Computational Graphs.} A computational graph illustrates a complex logical query through a graph, as shown in Figure \ref{fig:compute_graph}. Nodes symbolize the distribution of entities, and edges denote the transformations of this distribution. It helps to propagate queries through the KG. We can construct the graph from the query by converting the relation propagation to the projection operations, conjunction to the intersection of sets, negation to complement, and disjunction to the union of the entity distribution.}

{\setlength{\parindent}{0pt}\textbf{Rough Convex Cone.}
Here, we discuss the use of the rough convex cone to embed the MMKG.
\begin{definition}~\cite{zhang2021cone}
A subset $C \in \mathbb{R}^2$ is called a cone if, $\forall x \in C$, $\alpha \in \mathbb{R}_{\geq 0}$, $\alpha x \in C$. A cone is convex if, $\forall x_1, x_2 \in C$ and $\lambda_1, \lambda_2 \in \mathbb{R}_{\geq 0}$, we have $\lambda_1 x_1 + \lambda_2x_2 \in C$.
\end{definition}
Previous method proposed sector cone for embedding a (non-multi-modal) KG~\cite{zhang2021cone}. 
\begin{definition}
    A 2D closed cone is called a sector-cone, if its closure-complement or itself is convex.
\end{definition}
\begin{definition}
    The closure-complement of cone $C \subset \mathbb{R}^2$ is defined by $\Tilde{C} = cl(\mathbb{R}^2\setminus C)$, where $cl(.)$ is the closure of a set.
\end{definition}
To handle MMKGs, we extend the convex cones to rough convex cones.
\begin{definition}~\cite{liao2012rough}
    Let $W$ be a subspace of a space $S$, for $\alpha, \beta \in S$, $\alpha$ and $\beta$ are congruence with respect to $W$ ($\alpha \in R_w(\beta)$), if $\alpha - \beta \in W$. Here, $R_w(\beta)$ is the equivalence class containing the element $\beta$.
\end{definition}
\begin{definition}~\cite{liao2012rough}
    The rough approximation of a non empty answer set $X\subseteq S$, is defined as $R_wX = (\underline{R_w}X, \overline{R_w}X)$, where, 
\begin{subequations}
\begin{align}
    \underline{R_w}X &= \{x \in S \mid R_w(x) \subseteq X\}\\
    \overline{R_w}X &= \{x \in S \mid R_w(x) \cap X \neq \emptyset\}
\end{align}
\end{subequations}
$\underline{R_w}X$ and $\overline{R_w}X$ are lower and upper rough approximations of $X$, respectively. $X$ is called a $R_w$-rough convex cone if both $\underline{R_w}X$ and $\overline{R_w}X$ are convex cones. $R_wN X =  \overline{R_w}X - \underline{R_w}X$ is the fuzzy boundary region.
\end{definition}
Membership function in a rough set~\cite{pawlak1982rough} is defined as
\begin{subequations}
\begin{align}
    \mu_X^{R_w} (x) = 1, &\text{\hspace{20pt} iff } x \in \underline{R_w}X \label{eq:inrough}\\
    \mu_X^{R_w} (x) = 0, &\text{\hspace{20pt} iff } x \in S -\overline{R_w}X \label{eq:outrough} \\
    0 < \mu_X^{R_w} (x) < 1, &\text{\hspace{20pt} iff } x \in R_wNX \label{eq:midrough}
\end{align}
\end{subequations}
In MMKG, let all the entities $e \in \mathcal{V}$ are equivalent classes. For Q1, \emph{Toy Story (Poster)} $\in \overline{R_w}X$ since at least one sub-entity of the entity belongs to the answer set (\emph{green}), while \emph{Toy Story (Poster)} $\notin \underline{R_w}X$ as not all of its sub-entities belong to the answer set. Similarly, in Q2, \emph{John Lasseter} $\in \underline{R_w}X$, since the entire entity is an answer to the query.}

\begin{figure}[!t]
    \centering
    \includegraphics[width=0.27\paperwidth]{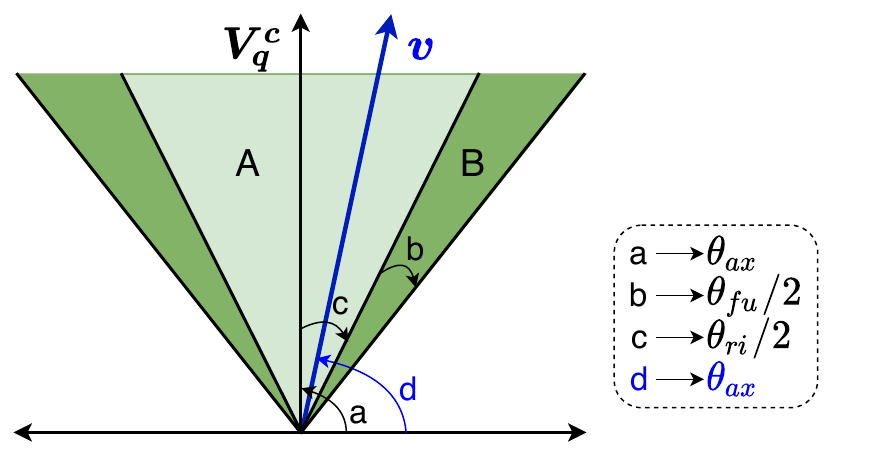}
    \caption{\model\ embedding for a query, $\boldsymbol{V_q^c}=(\theta_{ax},\theta_{ri},\theta_{fu})$, in $2$-dimension, with $\theta_{ax}$ being the angle between the co-ordinate axis and axis of the embedding. The region A (light green) across the axis is the rigid region ($\theta_{ri}$), and the region B (dark green) resembles the fuzzy (boundary) region ($\theta_{fu}$). The blue vector is the entity embedding ({\color{blue}$\boldsymbol{v}=(\theta_{ax},0, 0)$}).}
    \label{fig:fuzzyCone}
\end{figure}

{\setlength{\parindent}{0pt}\textbf{\model{} Representation.} The sector-cone is always axially symmetric, we can represent a $2$-dimensional sector-cone using a pair of parameters $(\theta_{ax}, \theta_{ri})$, where $\theta_{ax}$ is the axis and $\theta_{ri}$ is the aperture~\cite{zhang2021cone}. We use the idea of the rough convex cone, and present a two-dimensional \model{} as a triplet $(\theta_{ax}, \theta_{ri}, \theta_{fu})$ (Figure \ref{fig:fuzzyCone}). We have $\theta_{ri} \in [0,2\pi]$ as the rigid aperture, for lower approximation ($\underline{R_w}X$). $\theta_{fu} \in [0,2\pi-\theta_{ri}]$ is the extended fuzzy (boundary) region ($R_wN X$) along both sides of the rigid sector cone. $\theta_{ax} \in [-\pi, \pi)$ is an axis shared by both cones. Consider \model{} as two overlapping sector cones ($\underline{R_w}X$ and $\overline{R_w}X$) sharing a common semantic axis. All the points in the range $[\theta_{ax}-\theta_{ri}/2, \theta_{ax}+\theta_{ri}/2]$ would belong to the rigid sector cone, while all the points in $[\theta_{ax}-(\theta_{ri}+\theta_{fu})/2,\theta_{ax}-\theta_{ri}/2]\cup[\theta_{ax}+\theta_{ri}/2, \theta_{ax}+(\theta_{ri}+\theta_{fu})/2]$, would be in the fuzzy boundary sector cone.}

We can represent a \model{} as $R_0=(\theta_{ax}, \theta_{ri}, \theta_{fu}) \in \mathbb{K}$, where $\mathbb{K}$ is the space containing all triplets $(\theta_{ax}, \theta_{ri}, \theta_{fu})$. The $d$-ary Cartesian product of \model{} can be presented as 
\begin{equation}
    R=((\theta_{ax}^1, \theta_{ri}^1, \theta_{fu}^1), \ldots (\theta_{ax}^d, \theta_{ri}^d, \theta_{fu}^d)) \subset \mathbb{K}^d
\end{equation}
or $R=(\boldsymbol{\theta}_{ax}, \boldsymbol{\theta}_{ri}, \boldsymbol{\theta}_{fu})$, with $\boldsymbol{\theta}_{ax}=\{\theta_{ax}^1, \ldots, \theta_{ax}^d\} \in [-\pi, \pi)^d$, $\boldsymbol{\theta}_{ri}=\{\theta_{ri}^1, \ldots, \theta_{ri}^d\} \in [0, 2\pi]^d$, and $\boldsymbol{\theta}_{fu}=\{\theta_{fu}^1, \ldots, \theta_{fu}^d\} \in [0, 2\pi-\theta_{ri}]^d$.
\section{Method}\label{sec:method}
\begin{figure*}[!ht]
	\centering
	\includegraphics[width=0.8\paperwidth]{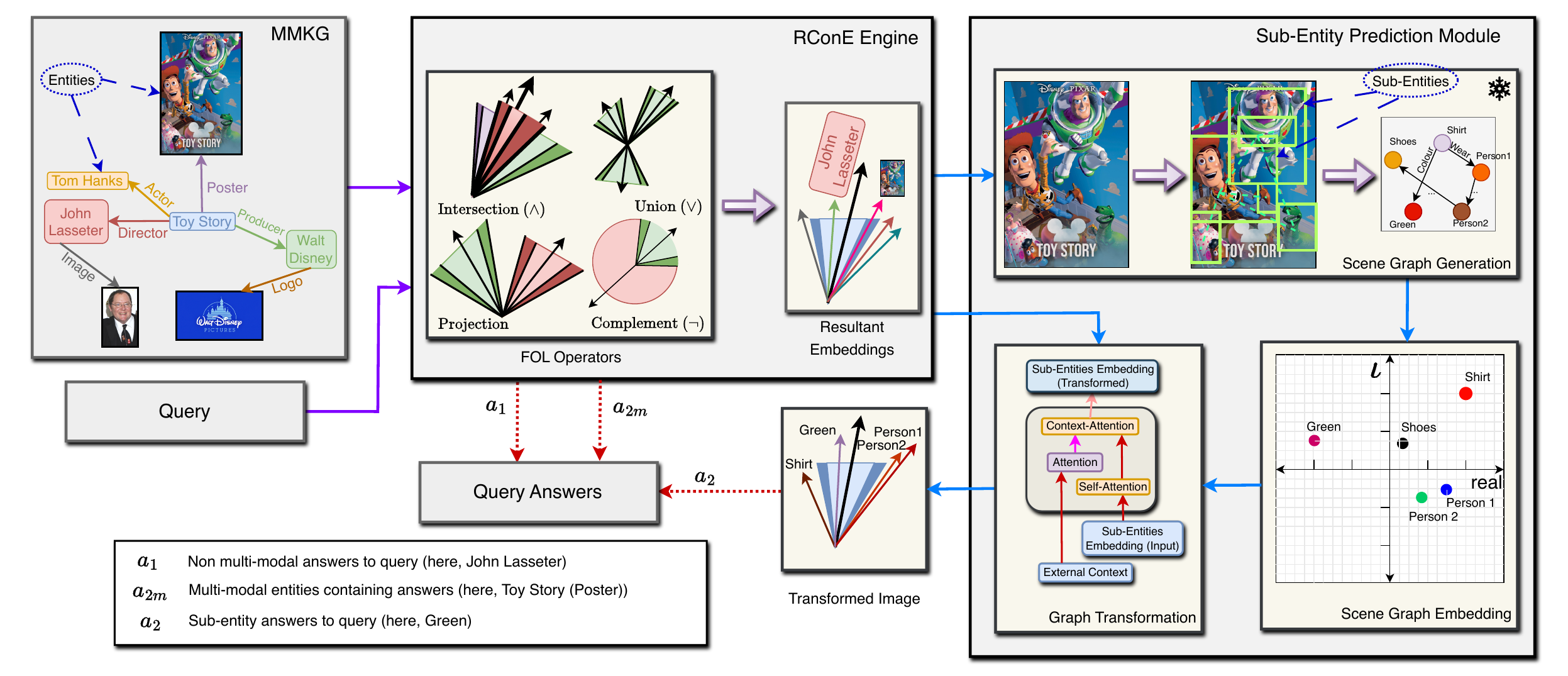}
	\caption{\textbf{Our proposed framework, \model{}:} The first sub-module, \submodel{} (RCE), embeds the input query and Multi-Modal Knowledge Graph (MMKG) to the embedding space (resultant embeddings). Then, the Sub-Entity Prediction module generates the sub-entities (fluorescent green boxes in Scene Graph Generation sub-module) and their embeddings from the candidate multi-modal entities in the fuzzy region (dark blue region in resultant embeddings), using three sub-modules -- Scene Graph Generation, Scene Graph Embedding and Graph Transformation.}
	\label{fig:model}
\end{figure*}

Figure \ref{fig:model} presents the flow diagram of our model \model{}. It comprises two modules, the \submodel{} (RCE) and the Sub-Entity Prediction module. First, the query and the MMKG are provided to the RCE. The RCE parses the query by following its computational graph (similar to Figure \ref{fig:compute_graph}) and processes all the FOL operators it encounters. The RCE outputs two entity sets – 1) Answer entities in the rigid region (\emph{John Lassester} in Q2). 2) Multi-modal entities consisting of the answers in the fuzzy region (\emph{Toy Story (Poster)} in Q1). Each candidate multi-modal entity in the fuzzy region (here, \emph{Toy Story (Poster)}) is passed through the Sub-Entity Prediction module. Using the Scene Graph Generation sub-module, the model detects sub-entities and relations among these sub-entities. It generates a sub-entity KG (scene graph) using these sub-entities and relations. Figure \ref{fig:mmkg} shows an example of the scene graph (the KG in the cloud). The generated scene graph is then embedded in the latent space using the Scene Graph Embedding module. Finally, the Graph Transformation module transforms the sub-entities to the \model’s embedding space. The transformation is such that the answer sub-entities (\emph{Green} in Q1) belong in the rigid region, and other sub-entities are embedded outside the \model{} embedding. In the following subsections, we describe each module in detail. A brief overview of the steps is presented in Algorithm 1 in the supplementary material.

\subsection{\submodel{} (RCE)}\label{sec:emb_mod}
As shown in Figure \ref{fig:model}, RCE takes MMKG and queries as input and embeds them in the \model{} embedding space. It traverses the query computational graph to generate the embedding. During traversal, RCE handles different FOL operators in a query. We first discuss the embedding procedure for queries and entities. Following it are the details about the transformation in embeddings based on the FOL operators. 

{\setlength{\parindent}{0pt}\textbf{Entity and Query Embedding.} 
Let $\mathcal{A}(q)$ be the entity set satisfying the query $q$. The embedding for $\mathcal{A}(q)$ is the cartesian product of $d$-ary (fuzzy-rigid) sector cones in $\mathbb{K}^d$ embedding space (Figure \ref{fig:fuzzyCone}). The embedding is presented as $\boldsymbol{V}^c_q=(\boldsymbol{\theta}_{ax},\boldsymbol{\theta}_{ri},\boldsymbol{\theta}_{fu})$, where $\boldsymbol{\theta}_{ax} \in [-\pi,\pi)^d$ is the semantic axis. $\boldsymbol{\theta}_{ri}\in[0,2\pi]^d$ is the rigid aperture enclosing the region $A$ around it. $\boldsymbol{\theta}_{fu} \in [0,2\pi-\theta_{ri}]^d$ is the fuzzy aperture enclosing region $B$ around the rigid cone.}

All entities belonging to the answer set (\emph{John Lasseter} in Q2) will have embeddings in the region $A$. The multi-modal entities in $G$, which have some part in the answer (\emph{Toy Story (Poster)} in Q1), will have embeddings in region $B$. 

The embedding of node $v \in \mathcal{V}_{\cup}$ is represented as $\boldsymbol{v}=(\boldsymbol{\theta}_{ax},\boldsymbol{0}, \boldsymbol{0})$. It can be considered as an answer set consisting of a single answer $v$. The embedding consists of no rigid and fuzzy areas (blue vector in Figure \ref{fig:fuzzyCone}).

The reasons we are using the rough sets (fuzzy (boundary) region) as an extension to the original ConE~\cite{zhang2021cone} model are
\begin{itemize}
    \item The multi-modal entities should operate in the same space as the unimodal entities as we are traversing the query in that space.
    \item The multi-modal entities containing the answer would be semantically similar to other answers. So, their embedding should also be closer. Hence, they would belong to the boundary region with partial membership (Equation \ref{eq:midrough}) to the answer set.
\end{itemize}

\begin{figure*}[!ht]
    \centering
    \subfloat[]{\includegraphics[width=0.243\textwidth]{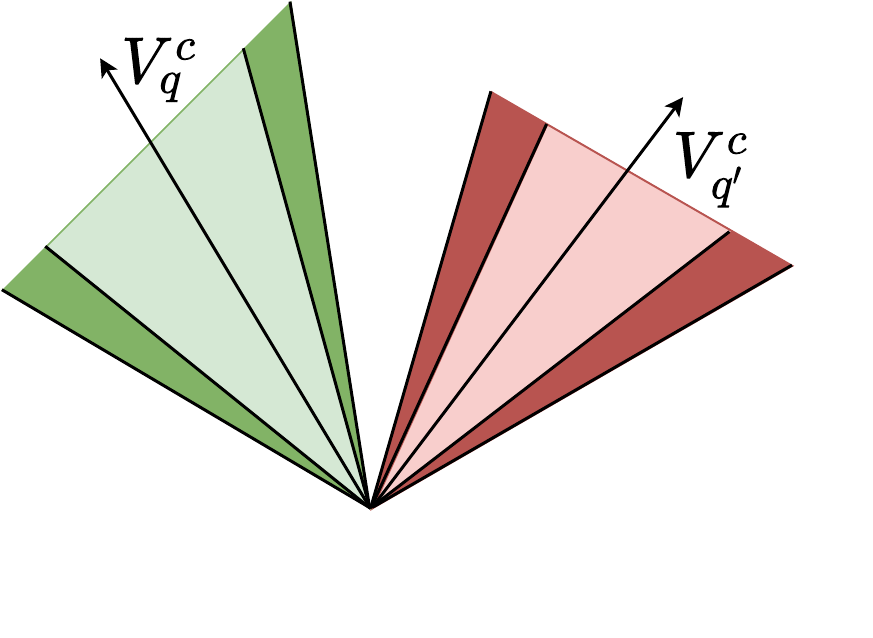}
    \label{fig:cone_projection}}
    \hfil
    \subfloat[]{\includegraphics[width=0.243\textwidth]{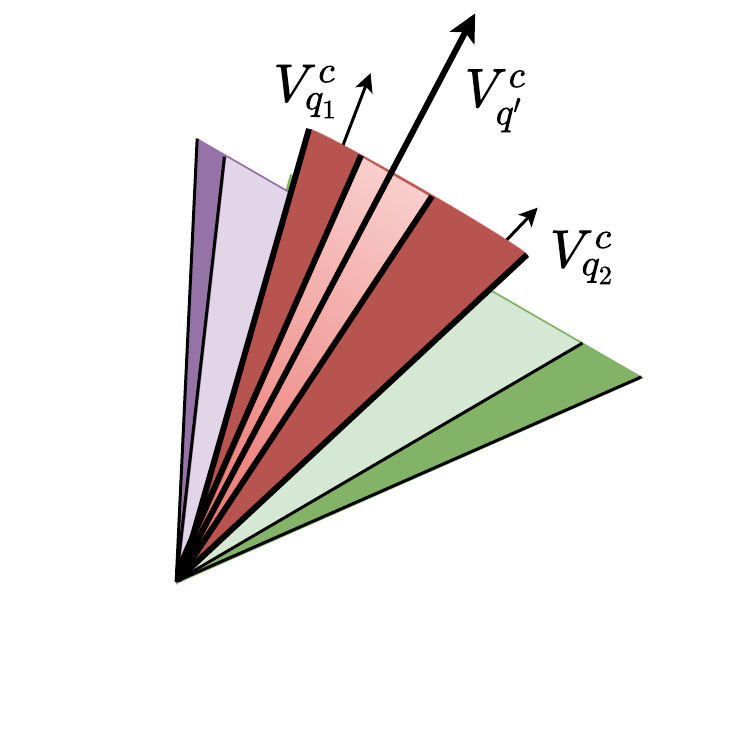}
    \label{fig:cone_intersection}}
    \hfil
    \subfloat[]{\includegraphics[width=0.243\textwidth]{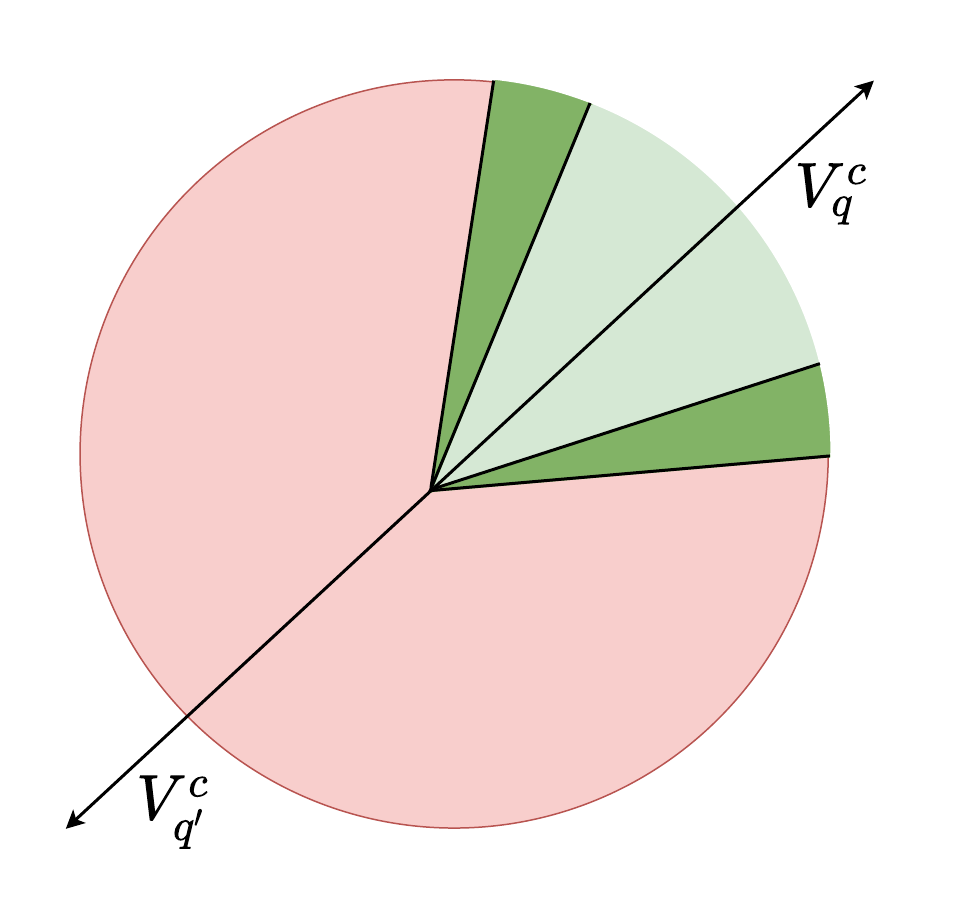}
    \label{fig:cone_complement}}
    \hfil
    \subfloat[]{\includegraphics[width=0.243\textwidth]{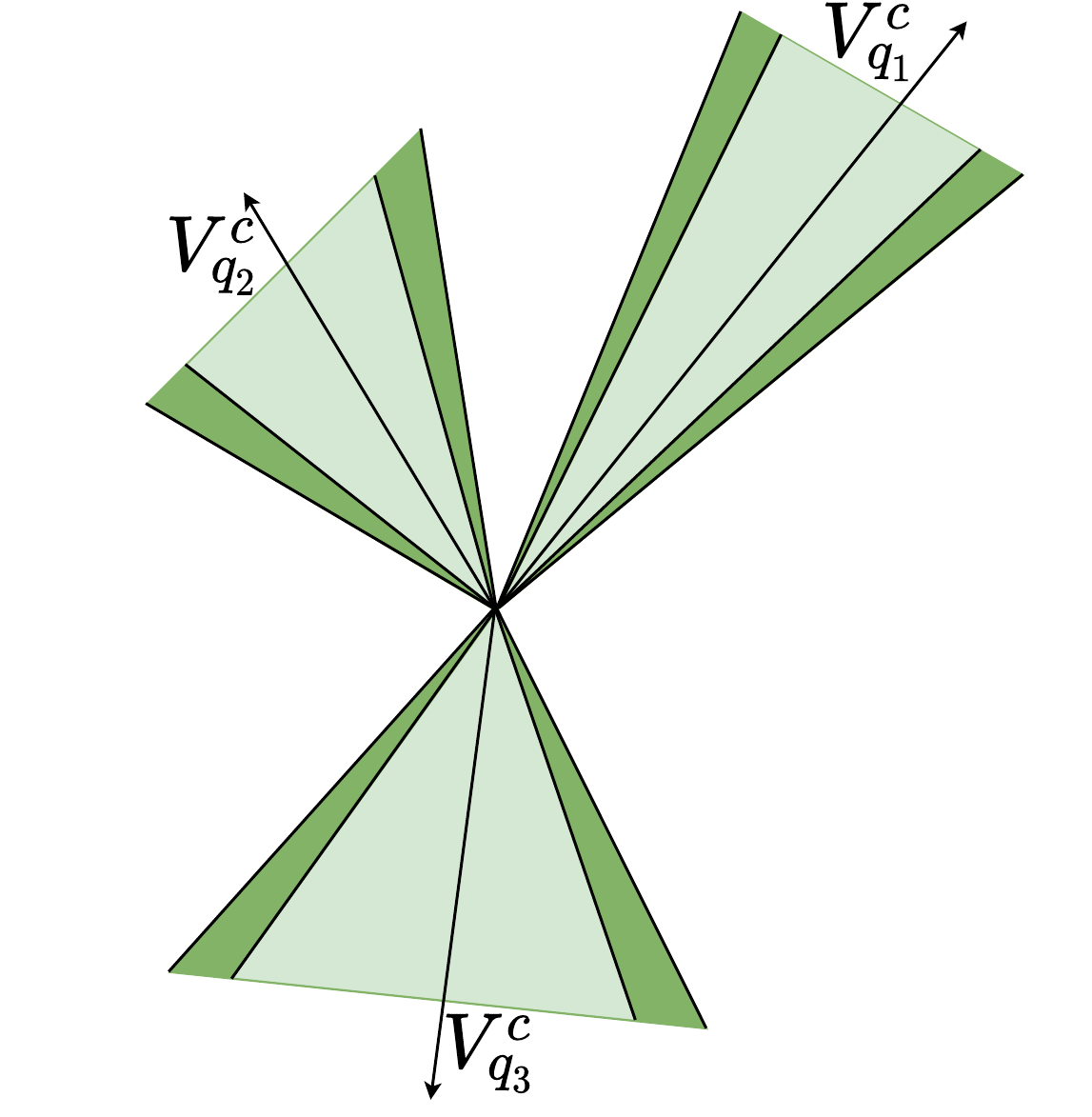}
    \label{fig:cone_union}}
    \caption{A representation of the impact of different logical operators in \submodel{} (RCE) ($1$-ary). Here (for (a) projection, (b) intersection, and (c) complement operators), $\boldsymbol{V}^{c}_{q_i}$ is the input \model{} embedding(s) to the operator, and $\boldsymbol{V}^{c}_{q'}$ is the output embedding. In the (d) union operator, the output is all the input embeddings ($\boldsymbol{V}^{c}_{q_i}$).}
    \label{fig:cone_operators}
\end{figure*}

{\setlength{\parindent}{0pt}\textbf{Logical Operators.}
The query computational graph consists of different logical operators (Intersection, Union, and Complement) along with the relation projection operator. The $\mathcal{A}(q)$'s \model{} embedding resembles the answer set that satisfies the processed query at any moment through this traversal. Figure \ref{fig:cone_operators} consists of the visual representation of how all the operations work on $1$-ary \model{} embedding. We present the modeling of each operator below.} Each operator has been extended to accomodate the fuzzy boundary part in ConE\cite{zhang2021cone}.

\textbf{Projection.}
Projection is a relation-dependent transformation of answer distribution ($\mathcal{A}(q)$), with the transformed embedding in the same space as the original one. We define projection mapping using the following function.
\begin{align}
    \mathcal{P}_r:\boldsymbol{V}^c_q \mapsto \boldsymbol{V}^{c}_{q'},  \;\;\; \boldsymbol{V}^c_q, \boldsymbol{V}^{c}_{q'} \in \mathbb{K}^d
\end{align}
We can consider projection as the shift in cones' angles based on the relation encountered. Given the relation embedding as $\boldsymbol{r}=(\boldsymbol{\theta}_{ax,r},\boldsymbol{\theta}_{ri,r},\boldsymbol{\theta}_{fu,r})$, the transformed cone would be dependent on $\boldsymbol{\theta}_{ax}+\boldsymbol{\theta}_{ax,r},\boldsymbol{\theta}_{ri}+\boldsymbol{\theta}_{ri,r}$ and $\boldsymbol{\theta}_{fu}+\boldsymbol{\theta}_{fu,r}$. We use an MLP of this tranformed cone to describe the projection, as presented below.
\begin{equation}\label{eqn:proj}
\resizebox{0.9\linewidth}{!}{$\mathcal{P}_r(\boldsymbol{V}_q) =\sigma_P(\boldsymbol{MLP}([\boldsymbol{\theta}_{ax}+\boldsymbol{\theta}_{ax,r};\boldsymbol{\theta}_{ri}+\boldsymbol{\theta}_{ri,r};\boldsymbol{\theta}_{fu}+\boldsymbol{\theta}_{fu,r}])) $}
\end{equation}
\begin{equation}
=[\boldsymbol{\theta'}_{ax};\boldsymbol{\theta'}_{ri};\boldsymbol{\theta'}_{fu}] \nonumber
\end{equation}
Here $[.;.;.]$ concatenates the three vectors, and $\sigma_{P}$ (\emph{Tanh}) keeps the output in the required range for all the $\theta$s (Equation \ref{eqn:proj_range}). 
\begin{equation}\label{eqn:proj_range}
\resizebox{\linewidth}{!}{$
	[\sigma_{P}(x)]_i = \begin{cases}
		\theta'^{i}_{ax} = \pi tanh(\lambda_1 x_i), \hfill \text{if, } i\leq d.\\
		\theta'^{i-d}_{ri} = \pi tanh(\lambda_2 x_i) +\pi, \hfill \text{if,	} i \in (d, 2d].\\
		\theta'^{i-2d}_{fu} = \pi tanh(\lambda_3x_i)+\pi -\theta'^{i-2d}_{ri}, \hfill \text{if, } i > 2d.
		      \end{cases}$}
\end{equation}
$[\sigma_{P}(x)]_i$ represents the $i$-th value of $\sigma_{P}(x)$. $\lambda_j$'s being the parameters to scale. Though \emph{Tanh} has a range $(-1,1)$, which may lead to never getting the boundary values. However, approximations due to hardware constraints can lead to these values as well.

\textbf{Intersection.}
\sloppy If we have $n$ query answer sets, denoted as $\mathcal{A}(q_1), \ldots, \mathcal{A}(q_n)$ with \model{} embeddings $\boldsymbol{V}_{q_1}^c, \ldots, \boldsymbol{V}_{q_n}^c$, respectively. We aim to identify the intersection query answer set $\mathcal{A}(q')$ by finding the shared region among these embeddings (Figure \ref{fig:cone_intersection}). The modified parameters can be expressed as.
\begin{align}
    \boldsymbol{\theta'}_{ax} &= SemanticAverage(\boldsymbol{V}_{q_1}^c, \ldots, \boldsymbol{V}_{q_n}^c) \nonumber\\
    \boldsymbol{\theta'}_{ri} &= RigidAverage(\boldsymbol{V}^c_{q_1}, \ldots, \boldsymbol{V}^c_{q_n}) \nonumber\\
    \boldsymbol{\theta'}_{fu} &= FuzzyAverage(\boldsymbol{V}^c_{q_1}, \ldots, \boldsymbol{V}^c_{q_n}) \nonumber
\end{align}
For $i$-th input embedding $\boldsymbol{V}^{c}_{q_i}$, let $\mathcal{B}_i$ be defined as
\begin{align}\label{eqn:semantic_inp}
    \mathcal{B}_i =& [\boldsymbol{\theta}_{ax}-\boldsymbol{\theta}_{ri}/2;\boldsymbol{\theta}_{ax}+\boldsymbol{\theta}_{ri}/2; \nonumber\\
    &\boldsymbol{\theta}_{ax}-(\boldsymbol{\theta}_{ri}+\boldsymbol{\theta}_{fu})/2;\boldsymbol{\theta}_{ax}+(\boldsymbol{\theta}_{ri}+\boldsymbol{\theta}_{fu})/2]_i
\end{align}

\textit{SemanticAverage.}
The final aperture $\boldsymbol{\theta'}_{ax}$ depends on the average of the intersecting cones' semantic centers. Similar to~\cite{zhang2021cone}, our method employs an attention module $[\boldsymbol{a}]$ to characterize the semantic center based on each input's rigid and fuzzy region. We utilize the relation below to calculate the attention mechanism.
\begin{equation}
    [\boldsymbol{a}]=\sigma_{s2}(W_{s2}\sigma_{s1}(W_{s1}\mathcal{B}))
\end{equation}
Here, $\sigma_{s1}$ and $\sigma_{s2}$ represent the ReLU and softmax functions, respectively. 

The resultant semantic center is determined by mapping the input axes to a unit circle, performing a weighted average of all the semantic centers (using the attention module), and reverting the points to their original form. The following relations describe this process.
\begin{align}
    [\boldsymbol{x};\boldsymbol{y}]=&\sum_{i=1}^n[\boldsymbol{a}_i\circ cos(\boldsymbol{\theta}_{i,ax});\boldsymbol{a}_i\circ sin(\boldsymbol{\theta}_{i,ax})]\\
    \boldsymbol{\theta'}_{ax}=&\sigma_s(\boldsymbol{x},\boldsymbol{y})
\end{align}
Here, $\sigma_s$ is used to transform all the values into the required range as,
\begin{equation}
	[\boldsymbol{\theta'}_{ax}]_i=\begin{cases}
		\arctan([\boldsymbol{y}]_i/[\boldsymbol{x}]_i)+\pi, & if [\boldsymbol{x}]_i < 0,[\boldsymbol{y}]_i>0,\\
		\arctan([\boldsymbol{y}]_i/[\boldsymbol{x}]_i)-\pi, & if [\boldsymbol{x}]_i<0,[\boldsymbol{y}]_i<0,\\
		\arctan([\boldsymbol{y}]_i/[\boldsymbol{x}]_i), & otherwise
	\end{cases}
\end{equation}

\textit{RigidAverage.} 
Let $\boldsymbol{V}_{q_X}^c$ and $\boldsymbol{V}_{q_Y}^c$ be two input answer set embeddings, with rough sets $R_w X$ and $R_w Y$, respectively. The rigid region of the resultant answer set from the intersection operation~\cite{pawlak1982rough} is defined as
\begin{equation}
    \underline{R_w} (X \cap Y) = \underline{R_w} X \cap \underline{R_w} Y
\end{equation}
where $\underline{R_w} X$ and $\underline{R_w} Y$ are the rigid regions of the input rough sets $R_w X$ and $R_w Y$, respectively. Hence, the resultant \model{} embedding's rigid aperture $\boldsymbol{\theta'}_{ri}$  should be the subset of all intersecting apertures of $\boldsymbol{V}^c_{q_1}, \ldots, \boldsymbol{V}^c_{q_n}$ presented as Equation \ref{eqn:rigidavg} (similar to \cite{zhang2021cone}).
\begin{equation}\label{eqn:rigidavg}
    \boldsymbol{\theta'}^i_{ri}=\min\{\boldsymbol{\theta}_{1,ri}^i,...,\boldsymbol{\theta}_{n,ri}^i\}.\sigma
([DeepSets(\{\boldsymbol{V}_{q_j}\}^n_{j=1})]_i)
\end{equation}
Here, deep sets (Equation \ref{eqn:deepsets1}) is an MLP to learn the intersecting answer sets' context.
\begin{equation}\label{eqn:deepsets1}
    MLP\left(\frac{1}{n}\sum_{j=1}^{n} MLP(\mathcal{B})\right)
\end{equation}

\textit{FuzzyAverage.}
Unlike the \textit{RigidAverage}, the resultant fuzzy borders depend on the intersection of fuzzy-fuzzy regions and the fuzzy-rigid regions both, for two or more input answer sets. Let $\boldsymbol{V}_{q_X}^c$ and $\boldsymbol{V}_{q_Y}^c$ be two input answer set embeddings, with rough sets $R_w X$ and $R_w Y$, respectively. The upper rough approximation of the resultant answer set from the intersection operation~\cite{pawlak1982rough} is defined as
\begin{equation}
    \overline{R_w} (X \cap Y) \leq \overline{R_w} X \cap \overline{R_w} Y
\end{equation}
where $\overline{R_w} X$ and $\overline{R_w} Y$ are the upper rough approximation of the input rough sets $R_w X$ and $R_w Y$, respectively. Since $\overline{R_w} X = R_w NX + \underline{R_w} X$, we can deduce the following relation for the output \model{} embedding,
\begin{equation}\label{eq:fuzzyinmin}(\boldsymbol{\theta'}^i_{fu}+\boldsymbol{\theta'}^i_{ri})\leq\min\{(\boldsymbol{\theta}_{1,fu}^i+\boldsymbol{\theta}_{1,ri}^i),...,(\boldsymbol{\theta}_{n,fu}^i+\boldsymbol{\theta}_{n,ri}^i)\}
\end{equation}
We opt for a trivial boundary as well that the resultant fuzzy area would not be greater than the maximum fuzzy region,
\begin{equation}\label{eq:fuzzyinmax}
    \boldsymbol{\theta'}^i_{fu}\leq\max\{\boldsymbol{\theta}_{1,fu}^i,...,\boldsymbol{\theta}_{n,fu}^i\}    
\end{equation}
From Equation \ref{eq:fuzzyinmin} and  \ref{eq:fuzzyinmax}, we can construct the following relation for the resultant fuzzy boundary,
\begin{equation}
\begin{aligned}
\boldsymbol{\theta'}^i_{fu}=\sigma([DeepSets]).\min\{\max\{\boldsymbol{\theta}_{1,fu}^i,...,\boldsymbol{\theta}_{n,fu}^i\},\\ 
    \min\{(\boldsymbol{\theta}_{1,fu}^i+\boldsymbol{\theta}_{1,ri}^i),...,(\boldsymbol{\theta}_{n,fu}^i+\boldsymbol{\theta}_{n,ri}^i)\}-\boldsymbol{\theta'}^i_{ri}\}
    \end{aligned}
\end{equation}
where $\boldsymbol{\theta'}^i_{ri}$ is the updated rigid aperture, and DeepSets here is similar to that described in the \textit{RigidAverage}.

\textbf{Complement.}
The semantic axis for the resultant embedding ($\boldsymbol{V}_{q'}^c$) from the complement operator will be in the opposite direction to the input embedding ($\boldsymbol{V}_{q}^c$) (Figure \ref{fig:cone_complement}). The new axis is described as 
\begin{equation}\label{eqn:neg_app}
	[\boldsymbol{\theta'}_{ax}]_i = \begin{cases}
		[\boldsymbol{\theta}_{ax}]_i-\pi, & \text{if }[\boldsymbol{\theta}_{ax}]_i \geq 0,\\
		[\boldsymbol{\theta}_{ax}]_i +\pi, & \text{if } [\boldsymbol{\theta}_{ax}]_i<0
	\end{cases}
\end{equation}
The rigid region's angle of $\boldsymbol{V}_{q'}^c$ ($\boldsymbol{\theta'}_{ri}$) will be all the $2\pi$ region except the area covered by the input rigid ($\boldsymbol{\theta}_{ri}$) and fuzzy sector cone ($\boldsymbol{\theta}_{fu}$), that is,
\begin{equation}
    [\boldsymbol{\theta'}_{ri}]_i = 2 \pi - ([\boldsymbol{\theta}_{ri}]_i+[\boldsymbol{\theta}_{fu}]_i)
\end{equation}
For complement operation, the fuzzy embedding would be identical for both $\boldsymbol{V}_{q}^c$ and $\boldsymbol{V}_{q'}^c$ (the dark green region in Figure \ref{fig:cone_complement}). For a multi-modal entity $e$ belonging to the fuzzy area, if membership to the input embedding, $\boldsymbol{V}_{q_X}^c$, is $\mu_{X}^{R_w} (e) = k$, with $0<k<1$. After complement operation, it would be $\mu_{X'}^{R_w} (e) = (1-k)$. So,
\begin{equation}
    [\boldsymbol{\theta'}_{fu}]_i = [\boldsymbol{\theta}_{fu}]_i
\end{equation}

\textbf{Union.}
Handling the union operator at the end makes the model scalable~\cite{ren2020beta}. We convert the queries to the DNF form for execution. Suppose $\boldsymbol{V}_{q_i}^{c}=(\boldsymbol{\theta}_{i,ax},\boldsymbol{\theta}_{i,ri},\boldsymbol{\theta}_{i,fu})$ is the $i$th input \model{} embeddings. The union of all the input embeddings~\cite{zhang2021cone} would be
\begin{equation}
    \boldsymbol{V}_{q}^c = \{\boldsymbol{V}_{q_1}^c, \ldots,\boldsymbol{V}_{q_n}^c\}
\end{equation}
it can also be represented as
\begin{equation}
\begin{aligned}
\resizebox{0.88\linewidth}{!}{$\boldsymbol{V}_{q}^c =
(\{(\boldsymbol{\theta}_{1,ax}^1,\boldsymbol{\theta}_{1,ri}^1,\boldsymbol{\theta}_{1,fu}^1), \ldots,(\boldsymbol{\theta}_{n,ax}^1,\boldsymbol{\theta}_{n,ri}^1,\boldsymbol{\theta}_{n,fu}^1)\},$} \\ \resizebox{0.88\linewidth}{!}{$\ldots,
    \{(\boldsymbol{\theta}_{1,ax}^d,\boldsymbol{\theta}_{1,ri}^d,\boldsymbol{\theta}_{1,fu}^d), \ldots,(\boldsymbol{\theta}_{n,ax}^d,\boldsymbol{\theta}_{n,ri}^d,\boldsymbol{\theta}_{n,fu}^d)\})$}
\end{aligned}
\end{equation}
\begin{figure}[!t]
	\centering
	\includegraphics[scale=0.6]{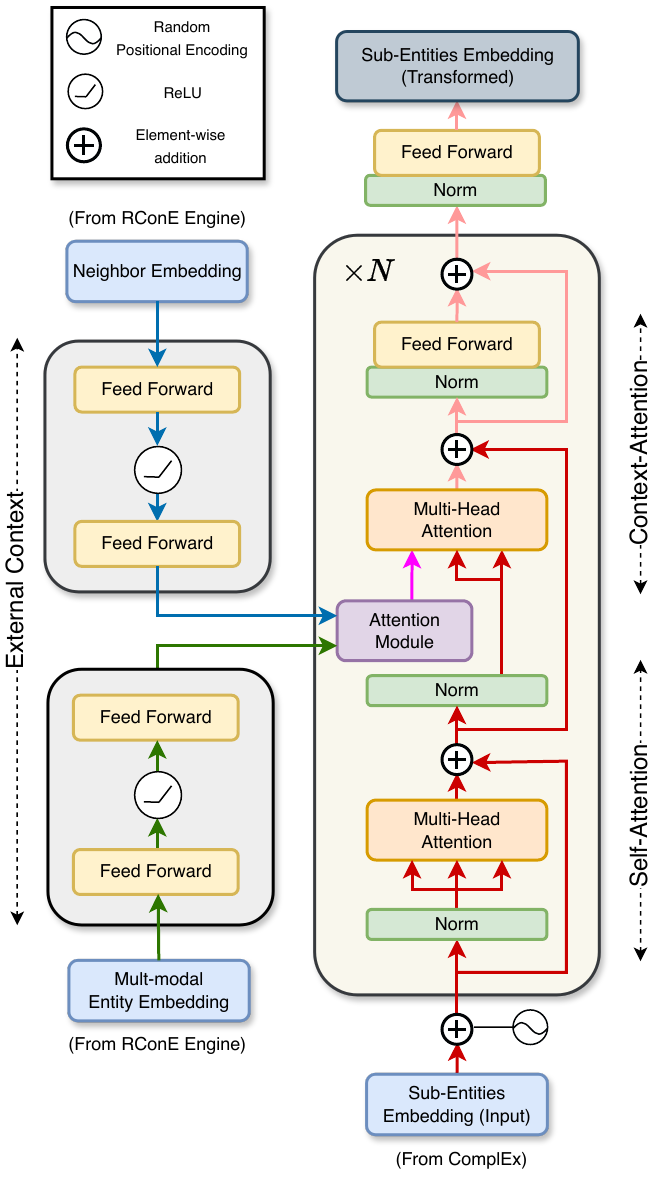}
	\caption{\textbf{Graph Transformation}: It transforms the sub-entities from the Scene Graph Embedding (ComplEx) space to the \model{} embedding space. It uses the multi-modal entity and its neighbor embedding from RCE in the attention module for external context. The input and output of the sub-module are highlighted in blue and grey, respectively.}
	\label{fig:transformer}
\end{figure}
\subsection{Sub-Entity Prediction}\label{sec:subentity_pred}
After passing through the \submodel{}, we receive entity sets belonging to the rigid region (\emph{John Lassester} in Q2) and fuzzy region (\emph{Toy Story (Poster)} in Q1). The Sub-Entity Prediction module takes candidate multi-modal entities from the fuzzy region (here, \emph{Toy Story (Poster)}) as input. It returns the embedding of the answer sub-entity \emph{Green} in the rigid region and other non-answer sub-entities outside the \model{} embedding. The module consists of three parts: Scene Graph Generation, Scene Graph Embedding, and Graph Transformation (Figure \ref{fig:model}). We present details of each sub-module below.

{\setlength{\parindent}{0pt}\textbf{Scene Graph Generation.}\label{sec:sgg} The sub-module creates a scene graph for each candidate multi-modal entity (here, \emph{Toy Story (Poster)}). The sub-model is trained on a large scene-graph dataset and is used in \model{} in a zero-shot manner to generate scene graphs. We use FCSGG~\cite{liu2021fully} as our scene graph generation algorithm.}

{\setlength{\parindent}{0pt}\textbf{Scene Graph Embedding.} This sub-module embeds the scene graph of each image received from the last sub-module in an embedding space. The embedding is based on the relations within that image to incorporate the context of the image. We use ComplEx~\cite{trouillon2016complex} to embed the scene graph.}

{\setlength{\parindent}{0pt}\textbf{Graph Transformation.} In the last step, we transform all the sub-entity embeddings of each scene graph embedding to the \model{} embedding space. The sub-module is based on the transformer~\cite{vaswani2017attention} (Figure \ref{fig:transformer}). In each layer, we have two multi-head attentions. The first attention is for the context of other sub-entities in the same multi-modal entity (each sub-entity of \emph{Toy Story (Poster)}). The second multi-head attention is for the context of the multi-modal entity and its n-hop neighbors (embedding of \emph{Toy Story (Poster)}, and its neighbors) from RCE. The attention~\cite{hamilton2017inductive} coefficients for external context are calculated in the attention module (Figure \ref{fig:transformer}) as}
\begin{equation}
    \resizebox{0.88\linewidth}{!}{$
    h_e^k \leftarrow \sigma (W . ( \{h^{k-1}_e \} \cup MEAN(\{h^{k-1}_u, \forall u \in N (\boldsymbol{v}_m)\})))$}
\end{equation}
\begin{equation}
    h_e^0 \leftarrow \boldsymbol{v}_m
\end{equation}
where $\boldsymbol{v}_m$ is the candidate multi-modal entity, and $N (\boldsymbol{v}_m)$ is the set of $\boldsymbol{v}_m$'s neighbor entities.
\subsection{Loss Function}\label{sec:loss_func}
\begin{table}[!t]
    \caption{Description of variables used in loss function.}
    \centering
     \resizebox{\columnwidth}{!}{ 
    \begin{tabular}{c|c}
    \toprule
    \textbf{Lower border} & \textbf{Upper border} \\
    \midrule
        $\boldsymbol{\theta}_{L_{ri}} = \boldsymbol{\theta}_{ax}-\boldsymbol{\theta}_{ri}/2$ &
        $\boldsymbol{\theta}_{U_{ri}}=\boldsymbol{\theta}_{ax}+\boldsymbol{\theta}_{ri}/2$\\
        $\boldsymbol{\theta}_{L_{fu}} = \boldsymbol{\theta}_{ax} - (\boldsymbol{\theta}_{ri}+\boldsymbol{\theta}_{fu})/2$ & $\boldsymbol{\theta}_{U_{fu}} = \boldsymbol{\theta}_{ax}+(\boldsymbol{\theta}_{ri}+\boldsymbol{\theta}_{fu})/2$\\
        $\boldsymbol{\theta}_{Lm_{fu}} = \boldsymbol{\theta}_{ax} - (\boldsymbol{\theta}_{ri}/2+\boldsymbol{\theta}_{fu}/4)$ & $\boldsymbol{\theta}_{Um_{fu}} = \boldsymbol{\theta}_{ax}+(\boldsymbol{\theta}_{ri}/2+\boldsymbol{\theta}_{fu}/4)$\\
    \bottomrule
    \end{tabular}}
    \label{tab:loss_formula}
\end{table}
There are two goals while constructing the loss function.
\begin{enumerate}
    \item For Type (I) answers, $e \in \mathcal{V}_j$ ($e_j \in \mathcal{V} \;|\;\gamma(e_j) = m_2$), we require the embedding of the nodes
    \begin{enumerate}
        \item $e_j$ (multi-modal entities) in the fuzzy region of \model{} (\emph{Toy Story (Poster)} in Q1).
        \item $e$ (the sub-entity answer) inside the rigid border of \model{} (\emph{Green} in Q1).
    \end{enumerate}
    \item For Type (II) answers, $e \in \mathcal{V}$, we require the embedding of the nodes within the rigid borders of \model{} (\emph{John Lasseter} in Q2).
\end{enumerate}
All other (sub) entities' embeddings in $\mathcal{V}$ and $\mathcal{V}_j$ should be outside \model. We define the loss function as
\begin{equation}\label{eqn:loss}
    L = L_c + \lambda_{l1} L_{se} + \lambda_{l2} L_{mm}
\end{equation}
Here, $L_c$ is the loss for Type (II) answers $e$ in $\mathcal{V}$ (goal 2), $L_{se}$ is for the sub-entity in {Type (I) answers $e$ in $\mathcal{V}_j$ (goal 1b), and $L_{mm}$ is for the multi-modal entities ($e_j$) in Type (I) answers (goal 1a). $\lambda_{l1}$ and $\lambda_{l2}$ are hyper-parameters. We use negative sampling loss in each case, as described by Equation \ref{eqn:loss1}.
\begin{equation}\label{eqn:loss1}
\begin{aligned}
    L_{neg}=&-\log \sigma (\gamma-d(\boldsymbol{v};\boldsymbol{V}^c_q)) -
\frac{1}{k}\sum_{i=1}^{k}\log\sigma(d(\boldsymbol{v'}_i;\boldsymbol{V}^c_q)-\gamma)
\end{aligned}
\end{equation}
Here $\boldsymbol{v}$ and $\boldsymbol{v'}$ are positive and negative samples, respectively. $d(\boldsymbol{v};\boldsymbol{V}_q^c)$ is the distance between the node $\boldsymbol{v}$ and query answer set embedding, $\boldsymbol{V}_q^c$ formulated as
\begin{equation}\label{eqn:distance_condis}
	d(\boldsymbol{v};\boldsymbol{V}^c_q) = \begin{cases}
		d_{con}(\boldsymbol{v};\boldsymbol{V}^c_q), & \text{if $q$ is conjunctive}\\
		d_{dis}(\boldsymbol{v};\boldsymbol{V}^c_q), & \text{if $q$ is disjunctive}
	\end{cases}
\end{equation}
\sloppy For a query in DNF, $q=q_1\vee\ldots \vee q_n$, the loss would be minimum amongst all the individual conjunctive distances, i.e., 
$d_{dis}(\boldsymbol{v};\boldsymbol{V}_q^c) = \min\{d_{con}(\boldsymbol{v};\boldsymbol{V}_{q_1}^c),...,d_{con}(\boldsymbol{v};\boldsymbol{V}_{q_n}^c)\}$.
The conjunctive distance $d_{con}(\boldsymbol{v};\boldsymbol{V}_q^c)$ consists of three parts as
\begin{equation}
\resizebox{0.88\linewidth}{!}{$
d_{con}(\boldsymbol{v};\boldsymbol{V}_q^c) = d_o(\boldsymbol{v};\boldsymbol{V}_q^c)+\lambda_1 d_i(\boldsymbol{v};\boldsymbol{V}_q^c)+\lambda_2
d_m(\boldsymbol{v};\boldsymbol{V}_q^c)$}
\end{equation}
where $d_i$ is the distance in the rigid region, $d_m$ is the distance in the fuzzy region, and $d_o$ is the distance outside \model. $\lambda_1$ and $\lambda_2$ are hyper-parameters. Let, $\boldsymbol{\theta}_{L_{ri}}$ and $\boldsymbol{\theta}_{U_{ri}}$ denote the lower and upper rigid borders, respectively (Table \ref{tab:loss_formula}). Similarly, let $\boldsymbol{\theta}_{L_{fu}}$, and $\boldsymbol{\theta}_{U_{fu}}$ be the fuzzy borders, respectively. Then $d_i$ and $d_m$ can be formalized as follows.
\resizebox{0.99\linewidth}{!}{\parbox{\linewidth}{%
\begin{subnumcases}{d_i =} 
||\min\{|sin((\boldsymbol{\theta}_{ax}^v-\boldsymbol{\theta}_{ax})/2)|,|sin(\boldsymbol{\theta}_{ri}/4)|\}||_1 \label{eqn:di_1}\\
    |||sin((\boldsymbol{\theta}_{ax}^v-\boldsymbol{\theta}_{ax})/2)|-|sin(\boldsymbol{\theta}_{ri}/4)|||_1 \label{eqn:di_2} 
\end{subnumcases}%
}}
\resizebox{\linewidth}{!}{\parbox{\linewidth}{%
\begin{subnumcases}{d_m =} 
||\min\{|sin((\boldsymbol{\theta}_{ax}^v-\boldsymbol{\theta}_{L_{ri}})/2)|,\nonumber\\|sin((\boldsymbol{\theta}_{ax}^v -\boldsymbol{\theta}_{U_{ri}})/2)|,|sin(\boldsymbol{\theta}_{fu}/4)|\}||_1\label{eqn:dm_1}\\
||\min\{|sin((\boldsymbol{\theta}_{ax}^v \boldsymbol{\theta}_{Lm_{fu}})/2)|,\nonumber\\|sin((\boldsymbol{\theta}_{ax}^v -\boldsymbol{\theta}_{Um_{fu}})/2)|,|sin(\boldsymbol{\theta}_{fu}/8)|\}||_1 \label{eqn:dm_2}
\end{subnumcases}%
}}
where, Equations \ref{eqn:di_1}, and \ref{eqn:dm_1} are for both goal 1b and 2; and Equations \ref{eqn:di_2}, and \ref{eqn:dm_2} are for the goal 1a.
The $\boldsymbol{\theta}_{Lm_{fu}}$ and $\boldsymbol{\theta}_{Um_{fu}}$ (Table \ref{tab:loss_formula}) are the centers of fuzzy regions on both the sides. The outside distance between \model{} and entity would be
\begin{equation}
\resizebox{\linewidth}{!}{$
d_o = ||\min\{|sin((\boldsymbol{\theta}_{ax}^v-\boldsymbol{\theta}_{L_{fu}})/2)|,|sin((\boldsymbol{\theta}_{ax}^v -
		\boldsymbol{\theta}_{U_{fu}})/2)|\}||_1 $}
\end{equation}
The loss functions for the scene graph generation and embedding are used as is~\cite{liu2021fully,trouillon2016complex}.

\section{Experiments}\label{sec:experiment}
\begin{table}[!t]
    \caption{Datasets description.}
    \centering
    \resizebox{0.45\textwidth}{!}{
    \begin{tabular}{l|ccc}
    \toprule
        \textbf{Name} & \textbf{Entities} & \textbf{Relations} & \textbf{Training Edges} \\
        \midrule
         FB15k & 14951 & 1345 & 483,142 \\ 
         FB15k-(237) & 14505 & 237 & 272,115 \\
         YAGO15k & 15283 & 32 & 122,886 \\
         DB15k & 14777 & 279 & 99,028 \\
        FB15k (NMM) & 14951 & 1345 & 483,142 \\ 
         \bottomrule
    \end{tabular}}
    \label{tab:datasets}
\end{table}
\subsection{Experimental Setup}
\begin{figure*}[!t]
    \centering
    \includegraphics[width=\textwidth]{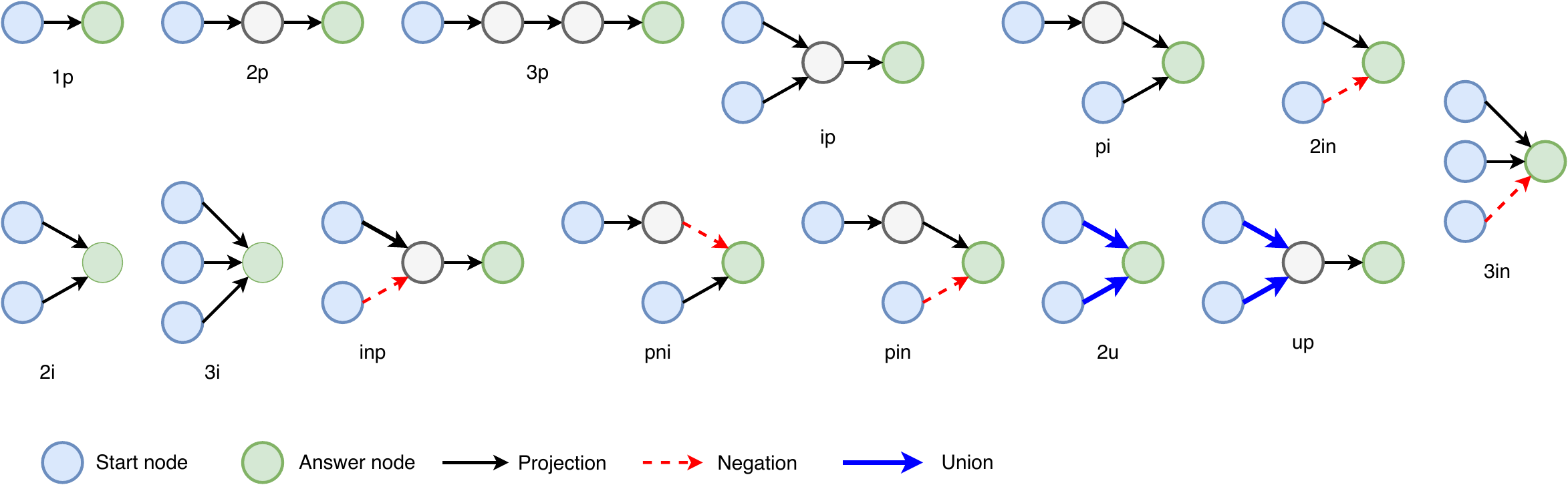}
    \caption{Query Structures used to construct queries for training \model{}. Here, \enquote*{p}, \enquote*{i}, \enquote*{n}, and \enquote*{u} are First Order Logical (FOL) operators projection, intersection, negation, and union, respectively.}
    \label{fig:query_structures}
\end{figure*}

{\setlength{\parindent}{0pt}\textbf{Datasets and Evaluation Metrics.} For evaluation, we use multi-modal datasets  FB15k, FB15k-(237), YAGO15k, and DB15k~\cite{liu2019mmkg}. We also train on FB15k (NMM)~\cite{bordes2013translating} (uni-modal dataset) to check whether the performance of our proposed model degrades compared to its non-multi-modal counterpart (ConE). Table \ref{tab:datasets} contains statistics of all the datasets. We use the average Mean Reciprocal Rank (MRR) and average HITS scores of all the answers to a query. We use these metrics for all three goals (1a, 1b, 2) described in Section \ref{sec:loss_func}.}

{\setlength{\parindent}{0pt}\textbf{Baselines.} Since \model{} is the first multi-modal logical query-answering algorithm, we choose non-multi-modal models, BetaE~\cite{ren2020beta}, ConE~\cite{zhang2021cone}, CylE~\cite{nguyen2023cyle}, and GNN-QE~\cite{zhu2022neural} as baselines. BetaE embeds the query answer distribution as a $d$-dimensional beta distribution. This was the first method to handle the negation ($\neg$) operator. ConE improved on BetaE and represented the answer distribution as $d$-ary sector-cones. CylE used three-dimensional shapes (cylinders) to add an extra dimension to embed the queries. GNN-QE uses NBF-Net~\cite{zhu2021neural} as a link prediction algorithm for projection operators and fuzzy logic for different operators. We cannot compare our model with path-based MMKG algorithms; hence, we omit them.}

{\setlength{\parindent}{0pt}\textbf{Query Generation.}\label{app:query_generation}
We propose a new query generation module that extends~\cite{ren2020beta} to get both Type (I) and Type (II) answers based queries from an MMKG. The model generates FOL queries with 14 query structures (\emph{1p}, \emph{2p}, \emph{3p}, \emph{2i}, \emph{3i}, \emph{pi}, \emph{ip}, \emph{2u}, \emph{up}, \emph{pni}, \emph{pin}, \emph{inp}, \emph{2in}, \emph{3in}) for both answer types (Figure \ref{fig:query_structures}). Here, \emph{p} is projection, \emph{i} is intersection, \emph{u} is union, and \emph{n} is negation.} The Query generation process is divided into three modules --  \emph{Scene Graph Generation}, \emph{Link Prediction}, and \emph{Query Traversal} (figure is presented in the supplementary material). 

\textbf{Scene Graph Generation.}
We select all the image entities from MMKG that are involved in the query generation process. The module generates a scene graph for each image using FCSGG~\cite{liu2021fully}. Note that the hyper-parameters/architecture here differs from Section \ref{sec:sgg} to keep our model independent.

\textbf{Link Prediction.}
Each generated scene graph is connected to the original MMKG using a KG embedding algorithm, ComplEx~\cite{trouillon2016complex}. ComplEx embeds sub-entities of all the scene graphs and the entities of MMKG into a lower-dimensional space. It then connects nodes from each scene graph to the MMKG based on the probability score. The output is a uni-modal KG.

\textbf{Query Traversal.}
We adopt the query generation method described in~\cite{ren2020beta} and incorporate Type I answers-based queries for all the 14 query structures (Figure \ref{fig:query_structures}) as well. Given a query structure, the algorithm chooses an answer randomly and backtracks to reach the source node while following the query structure. During traversal, it determines the relations randomly. It then executes the query again to get all the other entities that can be reached, given the source and the query.

\subsection{Hyper-Parameters and Model Selection}\label{app:hyperparameter}
\begin{table*}[!t]
    \caption{Average MRR (\%) score for candidate multi-modal entities (first five rows of each dataset) and answer sub-entities for Type (I) answers for \model{} with and without graph transformer (grey rows). Bold results indicate the best score.}
    \centering
    \begin{tabular}{l|c|ccc|cc|cc|cc|ccc|cc|c}
    \toprule
       \textbf{Dataset} & \textbf{Model} & \textbf{1p} & \textbf{2p} & \textbf{3p} & \textbf{2i} & \textbf{3i} & \textbf{pi} & \textbf{ip} & \textbf{2u} & \textbf{up} & \textbf{pni} & \textbf{pin} & \textbf{inp} & \textbf{2in} & \textbf{3in} & \textbf{AVG} \\
         \midrule
        \multirow{7}{*}{FB15k} & BetaE & 5.9 & 1.9 & 0.8 & 6.9 & 8.0 & 5.7 & 1.8 & 5.7 & 1.8 & 4.9 & 0.9 & 0.7 & 5.0 & 5.5 & 4.0 \\
        & ConE & 8.0 & 4.3 & 2.9 & 9.9 & 8.3 & 7.3 & 4.3 & 7.8 & 4.6 & 5.5 & 2.9 & 1.8 & 4.6 & 5.0 & 5.5 \\
        & CylE & 19 & 8.8 & 6.9 & 21 & 24 & 17 & 10 & 21 & 9.1 & 16 & 6.9 & 4.2 & 14 & 18 & 14 \\
        & GNN-QE & 17 & 4.7 & 1.1 & 23 & 34 & 19 & 9.3 & 16 & 4.6 & 11 & 2.7 & 3.5 & 10 & 15 & 12 \\
        & \model{} (1a) & \textbf{89} & \textbf{40} & \textbf{28} & \textbf{92} & \textbf{95} & \textbf{67} & \textbf{33} & \textbf{85} & \textbf{41} & \textbf{74} & \textbf{34} & \textbf{19} & \textbf{75} & \textbf{72} & \textbf{62} \\
        & \mycc \model{} (1b) & \mycc 72 & \mycc 66 & \mycc 56 & \mycc 63 & \mycc 64 & \mycc 66 & \mycc 75 & \mycc 75 & \mycc 68 & \mycc 56 & \mycc 59 & \mycc 72 & \mycc 65 & \mycc 51 & \mycc 65 \\
        & \mycc \model{} (1b) w/o trans & \mycc 50 & \mycc 53 & \mycc 46 & \mycc 51 & \mycc 57 & \mycc 56 & \mycc 59 & \mycc 58 & \mycc 53 & \mycc 47 & \mycc 47 & \mycc 54 & \mycc 54 & \mycc 45 & \mycc 52 \\
        \midrule
        \multirow{7}{*}{FB15k-(237)} & BetaE & 29 & 20 & 10 & 30 & 33 & 28 & 15 & 26 & 17 & 24 & 7.5 & 6.1 & 19 & 22 & 20 \\
        & ConE & 9.5 & 11 & 7.9 & 8.4 & 9.1 & 11 & 10 & 10 & 11 & 6.5 & 5.6 & 3.8 & 4.8 & 6.0 & 8.0 \\
        & CylE & 45 & 35 & 25 & 42 & 49 & 45 & 28 & 44 & 29 & 40 & 16 & 12 & 33 & 39 & 34 \\
        & GNN-QE & 10 & 2.5 & 0.4 & 18 & 18 & 7.7 & 2.0 & 11 & 1.7 & 7.9 & 1.1 & 1.0 & 5.6 & 9.2 & 6.9 \\
        & \model{} (1a) & \textbf{99} & \textbf{65} & \textbf{44} & \textbf{99} & \textbf{99} & \textbf{92} & \textbf{52} & \textbf{99} & \textbf{60} & \textbf{91} & \textbf{46} & \textbf{24} & \textbf{88} & \textbf{91} & \textbf{76} \\
        & \mycc \model{} (1b) & \mycc 77 & \mycc 62 & \mycc 59 & \mycc 70 & \mycc 71 & \mycc 69 & \mycc 71 & \mycc 84 & \mycc 64 & \mycc 48 & \mycc 58 & \mycc 65 & \mycc 57 & \mycc 49 & \mycc 65 \\
        & \mycc \model{} (1b) w/o trans & \mycc 59 & \mycc 53 & \mycc 50 & \mycc 56 & \mycc 61 & \mycc 57 & \mycc 60 & \mycc 65 & \mycc 52 & \mycc 44 & \mycc 48 & \mycc 52 & \mycc 48 & \mycc 44 & \mycc 54 \\
         \midrule
         \multirow{7}{*}{YAGO15k} & BetaE & 5.8 & 0.9 & 0.9 & 4.5 & 2.3 & 2.0 & 1.3 & 2.9 & 1.1 & 2.0 & 0.8 & 0.8 & 2.2 & 0.8 & 2.0 \\
         & ConE & 1.5 & 0.4 & 0.6 & 0.4 & 0.3 & 0.4 & 0.5 & 1.1 & 0.8 & 0.3 & 0.4 & 0.6 & 0.4 & 0.0 & 0.6 \\
        & CylE & 7.3 & 1.6 & 2.2 & 4.6 & 2.6 & 3.2 & 2.1 & 4.9 & 1.8 & 4.2 & 1.9 & 1.4 & 3.5 & 0.7 & 3.0 \\
         & GNN-QE & 8.9 & 0.6 & 0.5 & 13 & 6.0 & 4.0 & 3.0 & 7.7 & 1.1 & 3.3 & 0.8 & 1.0 & 2.9 & 2.0 & 3.9 \\
         & \model{} (1a) & \textbf{86} & \textbf{40} & \textbf{24} & \textbf{91} & \textbf{73} & \textbf{51} & \textbf{32} & \textbf{66} & \textbf{30} & \textbf{63} & \textbf{23} & \textbf{15} & \textbf{67} & \textbf{33} & \textbf{54} \\
         & \mycc \model{} (1b) & \mycc 49 & \mycc 47 & \mycc 46 & \mycc 54 & \mycc 50 & \mycc 44 & \mycc 47 & \mycc 49 & \mycc 45 & \mycc 44 & \mycc 42 & \mycc 44 & \mycc 47 & \mycc 44 & \mycc 47 \\
         & \mycc \model{} (1b) w/o trans & \mycc 39 & \mycc 40 & \mycc 39 & \mycc 45 & \mycc 41 & \mycc 35 & \mycc 43 & \mycc 41 & \mycc 40 & \mycc 36 & \mycc 35 & \mycc 38 & \mycc 39 & \mycc 33 & \mycc 39 \\
         \midrule
        \multirow{7}{*}{DB15k} & BetaE & 22 & 12 & 7.7 & 24 & 24 & 19 & 13 & 17 & 12 & 18 & 11 & 9.4 & 17 & 18 & 16 \\
        & ConE & 8.4 & 10 & 7.5 & 8.2 & 8.2 & 9.9 & 11 & 7.8 & 9.8 & 5.1 & 6.8 & 6.8 & 5.6 & 6.2 & 8.0 \\
        & CylE & 38 & 22 & 17 & 39 & 41 & 28 & 20 & 32 & 19 & 33 & 18 & 14 & 34 & 35 & 28 \\
        & GNN-QE & 23 & 6.9 & 1.4 & 41 & 50 & 23 & 8.8 & 25 & 4.2 & 15 & 2.3 & 2.9 & 10 & 25 & 17 \\
        & \model{} (1a) & \textbf{81} & \textbf{34} & \textbf{20} & \textbf{92} & \textbf{95} & \textbf{73} & \textbf{33} & \textbf{79} & \textbf{33} & \textbf{78} & \textbf{30} & \textbf{20} & \textbf{79} & \textbf{88} & \textbf{64} \\
        & \mycc \model{} (1b) & \mycc 44 & \mycc 40 & \mycc 42 & \mycc 43 & \mycc 44 & \mycc 39 & \mycc 40 & \mycc 47 & \mycc 41 & \mycc 40 & \mycc 40 & \mycc 40 & \mycc 39 & \mycc 41 & \mycc 41 \\
        & \mycc \model{} (1b) w/o trans & \mycc 40 & \mycc 37 & \mycc 38 & \mycc 45 & \mycc 45 & \mycc 38 & \mycc 39 & \mycc 45 & \mycc 38 & \mycc 37 & \mycc 33 & \mycc 35 & \mycc 35 & \mycc 36 & \mycc 39 \\
         \bottomrule
    \end{tabular}
    \label{tab:mm_results_multi-modal}
\end{table*}

{\setlength{\parindent}{0pt}\textbf{Scene Graph Generation.}
We choose FCSGG~\cite{liu2021fully} as a scene graph generation algorithm. The model consists of a Convolutional Neural Network for object detection and relation prediction in a single training module. The single training module results in substantial decrease in the inference time and number of parameters. This is critical, as the scene graph generation sub-module should not be a bottleneck and slow the overall training/inference of \model{}. Hence, it is a preferred choice compared to other algorithms. The model uses relation affinity fields to achieve better results on unseen visual relationships and helps transfer context between the objects detected and the relations predicted.}

The model implements zero-shot learning for custom image prediction and labels the entities and relations based on the training dataset. The assumption is that the training set would have enough entity and relation diversity to be useful for our dataset. We use Visual-Genome~\cite{krishna2017visual} for training.

The sub-module is used in both \model{} and query generation. To make them independent, we use HRNet-W48\cite{liu2021fully} for query generation and HRNet-W32 for training. During training, we compare the individual sub-entities in the two generated sub-graphs above, using GloVe embeddings~\cite{pennington2014glove}.

{\setlength{\parindent}{0pt}\textbf{Scene Graph Embedding.} We chose ComplEx as a translation-based model because of its relatively simpler architecture than the neural models. Like the scene graph generation sub-module, the simpler choice helps to avoid bottlenecks.} 

In the query generation module, we need the ComplEx algorithm to merge the scene graph generated for each multi-modal entity with the MMKG. We choose the threshold between [0, 1] to have at least one edge between the two graphs; we keep a cap of a maximum of 100 edges for each pair of graphs. The settings for the other hyper-parameters are the similar to ~\cite{trouillon2016complex}.

{\setlength{\parindent}{0pt}\textbf{Query Generation.}
We added a constraint for query generation that the query’s source node(s) would always belong to the MMKG. For multi-modal datasets, we include a fraction of the images (5\%) described in the original datasets to get a good mix of multi-modal and non-multi-modal queries. Due to computational constraints, we generated $30\%$ of training queries for which at least one answer of the query is the sub-entity. The query generation statistics for each dataset are provided in the supplementary material.}

The baselines are trained on the uni-modal queries, as the models cannot detect the sub-entities for the Type (I) answers. Moreover, these models cannot distinguish between multi-modal entities that are answers (Type (II) answers) and those containing the sub-entities as answers (Type (I) answers’ candidate entity).

To make the comparison fair for baselines, we modified the test set (for baselines) to reach the candidate multi-modal entities only as answers in the case of Type (I) answers (get only to the \emph{Toy Story (Poster)} as answers, and not \emph{Green}, in case of Q1). The test set for Type (II) answers is identical for \model{} and the baselines. 

{\setlength{\parindent}{0pt}\textbf{\model.}
We train our model for 200k epochs for each multi-modal dataset. $\lambda_{l1}=1$, $\lambda_{l2}=0.005$ for the loss function (Equation \ref{eqn:loss}). For distance weightage ($d_o$, $d_i$, $d_m$), $\lambda_1=0.02, \lambda_2=1$, in the case of non-multi-modal queries, and $\lambda_1=0.9, \lambda_2=0.02$ for multi-modal queries. For the projection operator, we use $d=2400$ for the multi-modal datasets. We use one layer of transformer for training. Other configurations are the same as the ConE algorithm. Experiments are executed on NVIDIA GeForce RTX 2080 Ti GPUs.}
\begin{table*}[!ht]
    \caption{Average HITS (\%) score for candidate multi-modal entities (@10) (first five rows of each dataset) and answer sub-entities (@5) for Type (I) answers for \model{} with and without graph transformer (grey rows). Bold results indicate the best score.}
    \centering
    \begin{tabular}{l|c|ccc|cc|cc|cc|ccc|cc|c}
    \toprule
       \textbf{Dataset} & \textbf{Model} & \textbf{1p} & \textbf{2p} & \textbf{3p} & \textbf{2i} & \textbf{3i} & \textbf{pi} & \textbf{ip} & \textbf{2u} & \textbf{up} & \textbf{pni} & \textbf{pin} & \textbf{inp} & \textbf{2in} & \textbf{3in} & \textbf{AVG} \\
         \midrule
        \multirow{7}{*}{FB15k} & BetaE & 12 & 3.9 & 1.4 & 16 & 18 & 11 & 3.6 & 12 & 3.2 & 10 & 1.9 & 1.3 & 10 & 12 & 8.5 \\
        & ConE & 16 & 7.8 & 5.9 & 22 & 18 & 14 & 8.8 & 15 & 8.9 & 10 & 6.1 & 3.6 & 8.3 & 10 & 11 \\
        & CylE & 35 & 17 & 13 & 36 & 44 & 33 & 19 & 37 & 16 & 28 & 13 & 9.0 & 25 & 33 & 25 \\
        & GNN-QE & 45 & 11 & 2.0 & 50 & 70 & 43 & 23 & 41 & 10 & 31 & 5.9 & 7.3 & 28 & 38 & 29 \\
        & \model{} (1a) & \textbf{98} & \textbf{64} & \textbf{47} & \textbf{99} & \textbf{99} & \textbf{87} & \textbf{54} & \textbf{93} & \textbf{63} & \textbf{95} & \textbf{61} & \textbf{37} & \textbf{97} & \textbf{93} & \textbf{81} \\
        & \mycc \model{} (1b) & \mycc 92 & \mycc 90 & \mycc 84 & \mycc 89 & \mycc 90 & \mycc 90 & \mycc 93 & \mycc 95 & \mycc 91 & \mycc 81 & \mycc 84 & \mycc 92 & \mycc 88 & \mycc 79 & \mycc 88 \\
        & \mycc \model{} (1b) w/o trans & \mycc 70 & \mycc 73 & \mycc 66 & \mycc 63 & \mycc 72 & \mycc 74 & \mycc 80 & \mycc 77 & \mycc 74 & \mycc 72 & \mycc 68 & \mycc 75 & \mycc 76 & \mycc 69 & \mycc 72 \\
        \midrule
        \multirow{7}{*}{FB15k-(237)} & BetaE & 53 & 35 & 21 & 55 & 57 & 50 & 27 & 47 & 29 & 42 & 15 & 11 & 37 & 42 & 37 \\
         & ConE & 18 & 22 & 15 & 18 & 17 & 21 & 19 & 18 & 20 & 12 & 10 & 8.5 & 10 & 11 & 16 \\
         & CylE & 76 & 59 & 44 & 76 & 82 & 74 & 47 & 72 & 50 & 69 & 32 & 25 & 62 & 69 & 60 \\
         & GNN-QE & 24 & 4.9 & 0.7 & 38 & 40 & 17 & 3.7 & 25 & 3.5 & 18 & 2.7 & 2.0 & 14 & 21 & 15 \\
         & \model{} (1a) & \textbf{99} & \textbf{85} & \textbf{65} & \textbf{100} & \textbf{100} & \textbf{99} & \textbf{74} & \textbf{99} & \textbf{78} & \textbf{99} & \textbf{71} & \textbf{46} & \textbf{99} & \textbf{99} & \textbf{89} \\
        & \mycc \model{} (1b) & \mycc 95 & \mycc 87 & \mycc 86 & \mycc 93 & \mycc 93 & \mycc 90 & \mycc 91 & \mycc 97 & \mycc 89 & \mycc 78 & \mycc 82 & \mycc 88 & \mycc 83 & \mycc 75 & \mycc 88 \\
        & \mycc \model{} (1b) w/o trans & \mycc 79 & \mycc 73 & \mycc 71 & \mycc 78 & \mycc 81 & \mycc 78 & \mycc 80 & \mycc 85 & \mycc 74 & \mycc 70 & \mycc 70 & \mycc 75 & \mycc 71 & \mycc 66 & \mycc 75 \\
         \midrule
        \multirow{7}{*}{YAGO15k} & BetaE & 11 & 1.5 & 2.0 & 9.7 & 5.4 & 4.8 & 2.6 & 7.3 & 2.1 & 4.7 & 2.0 & 1.5 & 5.1 & 1.7 & 4.4 \\
        & ConE & 2.9 & 0.7 & 1.3 & 1.2 & 0.8 & 0.6 & 1.1 & 2.2 & 1.8 & 0.5 & 1.0 & 1.0 & 0.7 & 0.0 & 1.1 \\
        & CylE & 15 & 3.2 & 4.1 & 9.3 & 6.4 & 6.5 & 5.3 & 11 & 4.3 & 9.4 & 4.5 & 3.1 & 7.0 & 1.7 & 6.5 \\
        & GNN-QE & 18 & 1.5 & 1.0 & 17 & 10 & 8.3 & 6.8 & 12 & 2.6 & 8.0 & 1.8 & 2.5 & 7.6 & 3.1 & 7.2 \\
       & \model{} (1a) & \textbf{93} & \textbf{56} & \textbf{38} & \textbf{99} & \textbf{86} & \textbf{65} & \textbf{49} & \textbf{82} & \textbf{45} & \textbf{87} & \textbf{37} & \textbf{28} & \textbf{90} & \textbf{38} & \textbf{71} \\
       & \mycc \model{} (1b) & \mycc 76 & \mycc 76 & \mycc 77 & \mycc 85 & \mycc 81 & \mycc 72 & \mycc 76 & \mycc 79 & \mycc 75 & \mycc 74 & \mycc 74 & \mycc 75 & \mycc 77 & \mycc 76 & \mycc 77 \\
       & \mycc \model{} (1b) w/o trans & \mycc 53 & \mycc 52 & \mycc 56 & \mycc 55 & \mycc 55 & \mycc 49 & \mycc 61 & \mycc 55 & \mycc 55 & \mycc 51 & \mycc 53 & \mycc 53 & \mycc 55 & \mycc 47 & \mycc 54 \\
       \midrule
        \multirow{7}{*}{DB15k} & BetaE & 42 & 24 & 14 & 44 & 44 & 39 & 24 & 33 & 21 & 35 & 21 & 17 & 34 & 35 & 31 \\
        & ConE & 16 & 21 & 15 & 16 & 14 & 20 & 20 & 15 & 19 & 10 & 12 & 13 & 11 & 12 & 15 \\
        & CylE & 64 & 42 & 32 & 66 & 68 & 53 & 41 & 54 & 35 & 57 & 33 & 27 & 58 & 59 & 49 \\
        & GNN-QE & 51 & 13 & 3.1 & 69 & 77 & 45 & 19 & 52 & 8.8 & 32 & 5.1 & 5.8 & 23 & 50 & 32 \\
       & \model{} (1a) & \textbf{93} & \textbf{58} & \textbf{37} & \textbf{99} & \textbf{99} & \textbf{90} & \textbf{55} & \textbf{89} & \textbf{54} & \textbf{91} & \textbf{51} & \textbf{36} & \textbf{91} & \textbf{97} & \textbf{80} \\
       & \mycc \model{} (1b) & \mycc 74 & \mycc 70 & \mycc 72 & \mycc 73 & \mycc 71 & \mycc 67 & \mycc 71 & \mycc 80 & \mycc 72 & \mycc 71 & \mycc 69 & \mycc 69 & \mycc 69 & \mycc 71 & \mycc 71 \\
       & \mycc \model{} (1b) w/o trans & \mycc 56 & \mycc 54 & \mycc 58 & \mycc 61 & \mycc 59 & \mycc 54 & \mycc 57 & \mycc 64 & \mycc 56 & \mycc 56 & \mycc 51 & \mycc 52 & \mycc 55 & \mycc 54 & \mycc 56 \\
       \bottomrule
    \end{tabular}
    \label{tab:mm_hits10_multi-modal}
\end{table*}
\begin{table*}[!t]
    \caption{Average MRR (\%) score for non-multi-modal query answers (Type (II) answers). Bold results indicate the best score.}
    \centering
    \begin{tabular}{l|c|ccc|cc|cc|cc|ccc|cc|c}
    \toprule
       \textbf{Dataset} & \textbf{Model} & \textbf{1p} & \textbf{2p} & \textbf{3p} & \textbf{2i} & \textbf{3i} & \textbf{pi} & \textbf{ip} & \textbf{2u} & \textbf{up} & \textbf{pni} & \textbf{pin} & \textbf{inp} & \textbf{2in} & \textbf{3in} & \textbf{AVG} \\
         \midrule
        \multirow{5}{*}{FB15k} & BetaE & 42 & 12 & 12 & 33 & 47 & 25 & 12 & 15 & 12 & 7.1 & 4.6 & 7.2 & 8.5 & 9.1 & 17 \\
        & ConE & 68 & 27 & 22 & 55 & 67 & 44 & 30 & 41 & 24 & 14 & 8.2 & 11 & 15 & 15 & 32 \\
        & CylE & 74 & 30 & 26 & 59 & 70 & 48 & 35 & 47 & 27 & 13 & 7.3 & 12 & 14 & 14 & 34 \\
        & GNN-QE & \textbf{84} & \textbf{58} & \textbf{43} & \textbf{77} & \textbf{83} & \textbf{69} & \textbf{59} & \textbf{68} & \textbf{49} & \textbf{32} & \textbf{30} & \textbf{35} & \textbf{43} & \textbf{42} & \textbf{55} \\
        & \model{} (2) & 57 & 23 & 21 & 46 & 59 & 37 & 26 & 25 & 21 & 10 & 6.6 & 10 & 11 & 12 & 26 \\
        \midrule
        \multirow{5}{*}{FB15k-(237)} & BetaE & 37 & 10 & 9.1 & 25 & 37 & 20 & 11 & 11 & 9.5 & 3.6 & 3.5 & 7.2 & 4.9 & 7.0 & 14 \\
        & ConE & 42 & 11 & 10 & 30 & 43 & 23 & 14 & \textbf{13} & 10 & 3.9 & 4.4 & 7.6 & 5.6 & 9.0 & 16 \\
        & CylE & \textbf{43} & \textbf{13} & \textbf{11} & 33 & 46 & 26 & 16 & \textbf{13} & 11 & 3.5 & 3.8 & 8.3 & 4.8 & 8.3 & 17 \\
        & GNN-QE & 40 & \textbf{13} & \textbf{11} & \textbf{38} & \textbf{53} & \textbf{30} & \textbf{19} & \textbf{13} & \textbf{12} & \textbf{5.4} & \textbf{6.3} & \textbf{9.2} & \textbf{6.7} & \textbf{15} & \textbf{19} \\
        & \model{} (2) & 39 & 11 & 10 & 28 & 40 & 21 & 13 & 11 & 9.9 & 3.9 & 3.6 & 7.7 & 5.8 & 8.7 & 15 \\
         \midrule
        \multirow{5}{*}{YAGO15k} & BetaE & 77 & 60 & 43 & 88 & 93 & 70 & 54 & 79 & 58 & 48 & 39 & 33 & 58 & 57 & 61 \\
        & ConE & 88 & 76 & 66 & 96 & 98 & 79 & 61 & 94 & 77 & 64 & 62 & 46 & 79 & 72 & 76 \\
        & CylE & 91 & 82 & 73 & 97 & 98 & 81 & 69 & 97 & 82 & 54 & 55 & 48 & 64 & 62 & 75 \\
        & GNN-QE & \textbf{99} & \textbf{93} & \textbf{82} & \textbf{100} & \textbf{99} & \textbf{99} & \textbf{98} & \textbf{99} & \textbf{96} & \textbf{98} & \textbf{94} & \textbf{93} & \textbf{99} & \textbf{99} & \textbf{96} \\
        & \model{} (2) & 77 & 69 & 62 & 90 & 94 & 70 & 55 & 76 & 68 & 54 & 47 & 39 & 63 & 59 & 66 \\
         \midrule
        \multirow{5}{*}{DB15k} & BetaE & 76 & 53 & 33 & 92 & 96 & 80 & 61 & 86 & 55 & 59 & 37 & 33 & 62 & 56 & 63 \\
        & ConE & 85 & 64 & 47 & 97 & 98 & 87 & 67 & 96 & 71 & 72 & 54 & 41 & 78 & 71 & 73 \\
        & CylE & 86 & 71 & 54 & 97 & 98 & 88 & 76 & 98 & 77 & 64 & 50 & 47 & 63 & 60 & 73 \\
        & GNN-QE & \textbf{99} & \textbf{83} & \textbf{68} & \textbf{99} & \textbf{99} & \textbf{97} & \textbf{91} & \textbf{99} & \textbf{84} & \textbf{98} & \textbf{88} & \textbf{86} & \textbf{99} & \textbf{99} & \textbf{92} \\
        & \model{} (2) & 74 & 56 & 40 & 92 & 96 & 77 & 60 & 80 & 59 & 61 & 39 & 32 & 64 & 54 & 63 \\
        \bottomrule
    \end{tabular}
    \label{tab:nmm_results_multi-modal}
\end{table*}

\subsection{Discussion}
[RQ1, RQ2] Logical query answers involving multi-modal entities and consisting of a sub-entity as an answer.

As discussed earlier, for an MMKG, we generate $30\%$ of queries consisting of at least one Type (I) answer and $70\%$ of queries with at least one Type (II) answer. Table \ref{tab:mm_results_multi-modal} consists of the average MRR score for all the Type (I) answers to queries on four multi-modal datasets. The table consists of two rows of results for our model, \model{} (1a) and \model{} (1b) represent scores for the respective goals presented in Section \ref{sec:loss_func}. The \model{} (1a) consists of the average MRR score for the candidate multi-modal entities (\emph{Toy Story (Poster)} in Q1), and \model{} (1b) (grey row) consists of the average MRR score for the answer sub-entities (\emph{Green} in Q1). As the baselines cannot detect the sub-entities for Type (I) answers, we calculate the scores for the candidate multi-modal entities. This comparison aims to evaluate if the baseline models can detect even these candidate entities and if their embedding is any closer to other answers. The first five rows in Table \ref{tab:mm_results_multi-modal} show the comparison (for these candidate entities). As can be seen in the table, the \model{} (1a) model performs significantly better than the baselines, with an average score improvement of $48$ in the case of FB15k, from the next best method (CylE). The MRR score decreases as we increase query complexity ($1p \rightarrow 2p \rightarrow 3p$). Though we cannot compare the sub-entity score (\model{} (1b)) with baselines, the values are significantly high to validate our model's performance. Table \ref{tab:mm_hits10_multi-modal} shows the average HITS@10 score for candidate multi-modal entities and the HITS@5 score for answer sub-entities for Type (I) answers. We choose the HITS@5 score rather than HITS@10 for the sub-entities in Table \ref{tab:mm_hits10_multi-modal} since the number of sub-entities in a scene graph is less when compared with entities in a graph. So, it would be more competitive to present the HITS@5 score rather than 10. The HITS score shows similar trends as the MRR score.

We compare the MRR score of our model \model{} for the Type (II) answers (\emph{John Lasseter} in Q2) as \model{} (2) for the goal 2 in Section \ref{sec:loss_func}. Since we train our model on both Type (I) and Type (II) answer type queries, this experiment tells the impact on the score for \model{} in the case of non-multi-modal answers (in MMKGs). As we can see from Table \ref{tab:nmm_results_multi-modal}, the MRR score of \model{} is less than its counterpart, ConE, in most cases. However, our model performs equivalent or better than BetaE based on average (AVG) scores for all the datasets. HITS scores with similar trends are presented in the supplementary material.

We can expect that the baseline models will perform better in Table \ref{tab:nmm_results_multi-modal} as these models are finetuned to handle only one type of query answer (Type (II) answers). In contrast, our model is more generalized and can handle both types of answers (Type (I) and Type (II)) simultaneously. \model{}’s decrease in the MRR score is reasonable, as it is still performing on par or better than BetaE on average. Moreover, the decline in Table \ref{tab:nmm_results_multi-modal} is much less than the significant gain \model{} gets for detecting the candidate entities in Table~\ref{tab:mm_results_multi-modal} (\model{} (1a)). Hence, overall, our model performs better than its counterpart (ConE) if we combine Table \ref{tab:mm_results_multi-modal} and \ref{tab:nmm_results_multi-modal} scores in the case of MRR score (similar trends for HITS score) for an MMKG dataset.

\subsection{Ablation Study}
We study the impact of adding the transformation module to \model{}. The model is trained without the transformation module. The results are presented as \emph{\model{} (1b) w/o trans} for the MRR and HITS score in Tables \ref{tab:mm_results_multi-modal} and \ref{tab:mm_hits10_multi-modal}, respectively. As we see in both the tables, the average performance degrades as we remove the sub-module (compared to \model{} (1b)). For the MRR score, the most difference in values is for the FB15k dataset (Table \ref{tab:mm_results_multi-modal}). RConE (1b) outperforms both Table \ref{tab:mm_results_multi-modal} and \ref{tab:mm_hits10_multi-modal} for individual query types except for the MRR scores of \emph{2i} and \emph{3i} in the DB15k dataset. The results validate the addition of the transformation module in our model.

\subsection{Fuzzy (Boundary) Region Integration Analysis}
In this sub-section, we check the impact of adding the fuzzy region to the original ConE embedding. For this, we compare the neural implementation of the projection operator with the actual set operator. However, the neural method may not precisely imitate the set operator. The goal is to check the performance using the fuzzy region compared to other components and baselines. We use the model trained on the FB15k dataset. 

We randomly generate $8000$ pairs of \model{} embeddings $(A_i, B_i)$, such that $A_i \subseteq B_i$. The goal is to check if, after the projection operation, the final embeddings $A_{Pi}$ and $B_{Pi}$ are still holding the relation $A_{Pi} \subseteq B_{Pi}$. We select an arbitrary relation, $r_i$, from the relation set in the FB15k dataset for the projection. We check the intersection area by calculating the score as $sc = Ar(A_{Pi} \cap B_{Pi})/Ar(A_{Pi})$, where $Ar()$ is the area of the sector cone. The score for the rigid region is $41\%$, while for the fuzzy region, it is $57\%$. The score for the ConE model is $46\%$. For the actual projection operator, the rigid region slightly drops in the score if we compare it with the baseline, and the score of the fuzzy part lies in the same range (better in this case). Hence, the fuzzy region performs on par with our model’s other neural structures and baselines. Therefore, the component addition is reasonable.

\subsection{Time Complexity Analysis}
[RQ3] Efficiency and scalability of \model{}.

{\setlength{\parindent}{0pt}\textbf{Efficiency.} Generating scene graphs for each multi-modal entity before training, even if there is no query related to it, will incur extra overhead. To counter this, in \model{}, we extract the scene graph for the multi-modal entities belonging to the fuzzy region only, which substantially decreases the computational cost.}

{\setlength{\parindent}{0pt}\textbf{Scalability.}
For the RCE module, we are adding a parameter, $\theta_{fu}$, in the original ConE architecture. Its space complexity depends on the number of relations in the MMKG, while its time complexity for each operator in RCE is on par with $\theta_{ri}$. Hence, it would not have much impact on the computation cost as we scale (compared to ConE). The Sub-Entity Prediction module uses three architectures: Scene Graph Generation, Scene Graph Embedding, and Graph Transformation. We train the Scene Graph Generation module to generate graphs to a dataset (Visual Genome) which is independent of the MMKG size. The Scene Graph embedding module's time complexity depends on the scene graph's size and is independent of the size of the KG. For the Graph Transformation module, we have neighbors of the multi-modal entity for external context. In real-world scale-free networks, the degree distribution follows near power law, with few nodes having a substantial number of neighbors, so it would have little impact as we scale. Hence, we can say that \model{} is scalable.}

\section{Conclusion and Future Work}\label{sec:futurework}
In this work, we proposed a novel method, \model{}, for extracting information from multi-modal entities to resolve logical queries. Our model is the first to answer logical queries related to multi-modal entities and can handle queries with sub-entities as answers. Our proposed model can also generalize well in handling multi-modal and non-multi-modal answer based queries. It performs very efficiently for multi-modal query answers, with minimal impact on non-multi-modal answers. We also provide a novel query generation module for MMKGs, which would help the research community generate queries with sub-entity answers.

The future goal is to explore the idea of inductive question answering and to get an answer by fusing information from multiple entities of different modalities for better context while answering a query. Apart from images, we will also consider working on other modalities for query answering.

\section*{Acknowledgments}
This work is partly supported by the University Grant Commission of India; the Infosys Center for AI, the Center of Design and New Media, and the Center of Excellence in Healthcare at IIIT-Delhi. The models were trained using Weights and Biases.

\bibliographystyle{IEEEtran}
\bibliography{sample-base}

\begin{IEEEbiographynophoto}{Mayank Kharbanda} received bachelor’s and master’s degrees in computer science from the University of Delhi, India, in 2018, and 2020, respectively. He is currently enrolled in a PhD program with the CSE department at IIIT-Delhi, India. 
\end{IEEEbiographynophoto}

\begin{IEEEbiographynophoto}{Rajiv Ratn Shah}
Rajiv Ratn Shah is an Associate Professor in the CSE department at IIIT-Delhi, India. He also serves as the head of the TCS Center of Design and New Media (CDNM), and MIDAS Research Lab. More information is available at \url{https://midas.iiitd.ac.in/}. 
\end{IEEEbiographynophoto}

\begin{IEEEbiographynophoto}{Raghava Mutharaju} is an Associate Professor in the CSE department of IIIT-Delhi, India and leads the Knowledgeable Computing and Reasoning (KRaCR) Lab. His research interests include knowledge graphs, ontology modelling, reasoning, querying, and its applications. More information is available at \url{https://kracr.iiitd.edu.in/}. 
\end{IEEEbiographynophoto}
\vfill

\newpage
\appendix

\subsection{Algorithm}\label{app:algorithm}
Algorithm \ref{algo:Fuzzycone} provides the pseudo-code for \model. We first pre-train the Scene Graph Generation module (Section IV-B) and initialize embeddings for all the entities and relations (Section IV-A Entity and Query Embedding) (Lines 1-2). For each training iteration, the model selects a random query $q_i \in Q_{train}$ (Figure  7) (Lines 3-4). The \submodel{} traverses the MMKG for the query $q_i$ according to the query's computational graph. It executes FOL operators until the end of the computational graph (Section IV-A Logical Operators) (Lines 5-7). This gives $\boldsymbol{V}^c_{q_i}$ as query embedding, an answer set for Type II answers (\emph{John Lasseter}), and candidate multi-modal entities (\emph{Toy Story (Poster)}). We then calculate the loss $L_c$, $L_{mm}$ (Equation 31) (Line 9). Following it, in the Sub-Entity Prediction module (Figure 4), the model generates a scene graph (Section IV-B Scene Graph Generation) for each candidate multi-modal entity containing the answer (\emph{Toy Story (Poster)}). It then embeds the scene graph and transforms it to the query embedding space (Section IV-B Scene Graph Embedding and Graph Transformation, respectively) (Lines 11-13). Loss $L_{se}$ is calculated for the sub-entity answer (Equation 31) (Line 15). The algorithm is continued till we reach the total number of iterations.

\subsection{Dataset Description}\label{app:dataset}
$\blacksquare$ \textbf{Freebase-15k (FB15k (NMM))}~\cite{bordes2013translating} is a subset of the original Freebase dataset of general facts. The dataset consists of 14,951 entities and 1,345 relations. Both entities and relations have at least 100 mentions. $\blacksquare$ \textbf{Freebase-15k Multi-Modal (FB15k)}~\cite{liu2019mmkg} is a multi-modal extension of the original FB15k (NMM) dataset. It consists, on average, 55.8 scaled images for each entity present in the KG. $\blacksquare$ \textbf{Freebase-15k (237) Multi-Modal (FB15k-(237))}~\cite{liu2019mmkg} is the FB15k dataset with 237 relations in it. The original non-multi-modal dataset was proposed to counter the problem of inverse relation test leakage~\cite{toutanova2015representing}. $\blacksquare$ \textbf{YAGO Multi-Modal (YAGO15k)}~\cite{liu2019mmkg} Similar to FB15k, this is a multi-modal extension of Yet Another Great Ontology (YAGO) dataset~\cite{fabian2007yago}. The YAGO dataset extends the information of Wordnet by merging general knowledge from Wikipedia. $\blacksquare$ \textbf{DBPedia Multi-Modal (DB15k)}~\cite{liu2019mmkg} DB15k is an extension of the original DBpedia dataset~\cite{auer2007dbpedia}, with images as entities included in it.

\subsection{Query Generation}
Figure \ref{fig:logical_query} represents the query generation module, and the method is presented in Algorithm \ref{algo:query_gen}. Table \ref{tab:query_count_stats} consists of the statistics for each query type for all the datasets. In algorithm \ref{algo:query_gen}, we first generate a scene graph for each image, which will be used for query generation (Lines 1-3). We embed the MMKG and all the scene graphs generated in the last step using the ComplEx algorithm. We do link prediction over the embedded graphs to connect each generated scene graph to the MMKG (Line 4). We generate queries for each query structure presented in Figure 7 using backtracking from an answer as described in ~\cite{ren2020beta} (Lines 5-7).

\subsection{Results}
Table \ref{tab:nmm_hits10_multi-modal} shows the average HITS score for Type (II) answers on all the multi-modal datasets. Similar to the MRR score, the HITS score is consistently better than BetaE, except for the DB15k dataset, where the drop is not substantial. Table \ref{tab:mm_hits3_multi-modal} shows the average HITS@3 score for both candidate and answer sub-entities. Table \ref{tab:nmm_hits3_multi-modal} shows the average HITS@3 score for Type (II) answers on all the multi-modal datasets. The trends in both tables are similar to the HITS@10 scores.

We also performed evaluations using the non-multi-modal dataset FB15k (NMM) to check the impact of changes made to \model{} with respect to the base ConE model, the MRR score is shown in Table \ref{tab:mrr_nmulti-modal}. As can be seen from the results, the performance does not drop from the original model.
\subsection{Runtime and Memory Efficiency Analysis}
As mentioned in the Introduction section, for large MMKGs, the pre-processing task of converting them into a non-multi-modal graph would incur high computational and space overhead. It consists of generating and storing the scene graph for each image. Let’s say there are $n$ image entities with, on average, $k$ edges in the scene graph of each image. If we are training on $t$-image nodes, $t<n$. Then, in the pre-processing task, we save about ($(n-t)*O(\text{SGG}+\text{SGE})$) for the computational cost, where SGG and SGE are time consumption for Scene Graph Generation and Scene Graph Embedding, respectively. We also save around $(n-t)*O(k+c)$ for the space cost, where $c$ is the average number of connections between a scene graph and the main graph, which will be significant at scale.
\begin{table*}[!t]
\caption{Query generation details for each query structure.}
\label{tab:query_count_stats}
    \centering
    \begin{tabular}{l|c|c|c|c}
    \toprule
         \textbf{Queries} & \multicolumn{2}{|c|}{\textbf{Training}} & \multicolumn{2}{|c}{\textbf{Test}} \\
         \midrule
         \textbf{Dataset} & \textbf{1p/2p/3p/2i/3i} & \textbf{2in/3in/inp/pin/pni} & \textbf{1p} & \textbf{others} \\
         \midrule
         FB15k & 273,710 & 27,371 & 68,549 & 8,000 \\
         FB15k-(237) & 149,689 & 14,698  & 29,962 & 8,000 \\
         YAGO15k & 107,982 & 10,798 & 75,835 & 4,000 \\
         DB15k & 107,982 & 10,798 & 29,799 & 4,000 \\
         FB15k (NMM) & 273,710 & 27,371 & 67,016 & 8,000 \\
         \bottomrule
    \end{tabular}
\end{table*}
\begin{table*}[!ht]
    \caption{Average HITS@10 (\%) score for non-multi-modal query answers (Type (II) answers). Bold results indicate the best score.}
    \centering
    \begin{tabular}{l|c|ccc|cc|cc|cc|ccc|cc|c}
    \toprule
       \textbf{Dataset} & \textbf{Model} & \textbf{1p} & \textbf{2p} & \textbf{3p} & \textbf{2i} & \textbf{3i} & \textbf{pi} & \textbf{ip} & \textbf{2u} & \textbf{up} & \textbf{pni} & \textbf{pin} & \textbf{inp} & \textbf{2in} & \textbf{3in} & \textbf{AVG} \\
         \midrule
        \multirow{5}{*}{FB15k} & BetaE & 69 & 25 & 23 & 57 & 72 & 45 & 25 & 32 & 24 & 15 & 10 & 15 & 18 & 20 & 32 \\
        & ConE & 89 & 44 & 37 & 78 & 87 & 65 & 47 & 61 & 42 & 28 & 18 & 23 & 32 & 33 & 49 \\
        & CylE & \textbf{91} & 49 & 42 & 80 & 89 & 69 & 54 & 69 & 45 & 27 & 16 & 24 & 30 & 31 & 51 \\
        & GNN-QE & 89 & \textbf{68} & \textbf{56} & \textbf{86} & \textbf{91} & \textbf{80} & \textbf{69} & \textbf{76} & \textbf{61} & \textbf{54} & \textbf{49} & \textbf{53} & \textbf{64} & \textbf{64} & \textbf{69} \\
        & \model{} (2) & 79 & 40 & 36 & 70 & 81 & 57 & 43 & 42 & 37 & 22 & 14 & 21 & 25 & 26 & 42 \\
        \midrule
        \multirow{5}{*}{FB15k-(237)} & BetaE & 58 & 20 & 17 & 46 & 59 & 36 & 22 & 22 & 19 & 8.0 & 7.5 & 15 & 11 & 15 & 25 \\
        & ConE & 62 & 23 & 19 & 51 & 65 & 40 & 25 & 25 & 20 & 8.5 & 9.8 & 16 & 12 & 19 & 28 \\
        & CylE & \textbf{64} & \textbf{24} & \textbf{21} & 53 & 67 & 43 & 28 & \textbf{26} & 22 & 7.4 & 8.3 & 17 & 11 & 17 & 29 \\
        & GNN-QE & 61 & \textbf{24} & 20 & \textbf{58} & \textbf{71} & \textbf{48} & \textbf{30} & 24 & \textbf{23} & \textbf{11} & \textbf{13} & \textbf{19} & \textbf{15} & \textbf{32} & \textbf{32} \\
        & \model{} (2)  & 59 & 22 & 19 & 49 & 62 & 37 & 25 & 22 & 19 & 8.5 & 7.9 & 16 & 12 & 19 & 27 \\
         \midrule
        \multirow{5}{*}{YAGO15k} & BetaE & 94 & 85 & 67 & 98 & 99 & 86 & 71 & 96 & 77 & 83 & 72 & 57 & 93 & 93 & 84 \\
        & ConE & 97 & 92 & 83 & 99 & 99 & 89 & 76 & 99 & 87 & 89 & 84 & 71 & 97 & 97 & 90 \\
        & CylE & 98 & \textbf{95} & \textbf{88} & 99 & 99 & 91 & 81 & 99 & 90 & 87 & 87 & 75 & 96 & 96 & 92 \\
        & GNN-QE & \textbf{100} & 94 & 84 & \textbf{100} & \textbf{100} & \textbf{99} & \textbf{98} & \textbf{100} & \textbf{97} & \textbf{99} & \textbf{95} & \textbf{95} & \textbf{99} & \textbf{99} & \textbf{97} \\
        & \model{} (2) & 90 & 88 & 78 & 98 & 99 & 84 & 69 & 86 & 81 & 83 & 75 & 63 & 92 & 90 & 84 \\
         \midrule
        \multirow{5}{*}{DB15k} & BetaE & 93 & 75 & 54 & \textbf{99} & \textbf{99} & 92 & 80 & 97 & 75 & 93 & 67 & 57 & 95 & 92 & 84 \\
        & ConE & 95 & 83 & 64 & \textbf{99} & \textbf{99} & 95 & 84 & \textbf{99} & 86 & 96 & 78 & 68 & 98 & 96 & 89 \\
        & CylE & 96 & \textbf{87} & 71 & \textbf{99} & \textbf{99} & 96 & 89 & \textbf{99} & \textbf{90} & 96 & 83 & 75 & 97 & 96 & 91 \\
        & GNN-QE & \textbf{99} & 86 & \textbf{75} & \textbf{99} & \textbf{99} & \textbf{98} & \textbf{92} & \textbf{99} & 89 & \textbf{99} & \textbf{92} & \textbf{91} & \textbf{99} & \textbf{99} & \textbf{94} \\
        & \model{} (2) & 89 & 77 & 58 & 98 & \textbf{99} & 90 & 78 & 90 & 78 & 91 & 65 & 56 & 93 & 87 & 82 \\
       \bottomrule
    \end{tabular}
    \label{tab:nmm_hits10_multi-modal}
\end{table*}
\begin{algorithm}[!t]
\textbf{Input}: MMKG - $G(\mathcal{V},\mathcal{R},\mathcal{U}, \mathcal{M})$, Training Query Set - $Q_{train}$ \\
\textbf{Output}: Trained model
\begin{algorithmic}[1]
\STATE $\text{Pre-train scene graph generation module (Section IV-B)}$

\STATE $\text{Initialize entity and relation embedding in }G \text{ as }$ $\boldsymbol{v}=(\boldsymbol{\theta}_{ax},\boldsymbol{0}, \boldsymbol{0})$, $\boldsymbol{r}=(\boldsymbol{\theta}_{ax,r},\boldsymbol{\theta}_{ri,r},\boldsymbol{\theta}_{fu,r})$\\

\WHILE{iterations $<$ total number of training iterations}
\STATE $\text{Randomly select } q_i \in Q_{train}$

\STATE $\text{Start from }\boldsymbol{v}_{start} \in q_i$

\WHILE{we reach end of computational graph}
\STATE Do one of the operation from projection, intersection, negation and union over query embedding. ($\boldsymbol{V}^c_q=(\boldsymbol{\theta}_{ax},\boldsymbol{\theta}_{ri},\boldsymbol{\theta}_{fu})$)
    (Section IV-A)
\ENDWHILE
\STATE Calculate $L_c$ and $L_{mm}$ (Equation 31).\\
\FOR{$v \in$ candidate multi-modal set}
\STATE $\text{generate scene graph(v)}$ 
\STATE $\text{embed scene graph(v)}$
\STATE $\text{transform scene graph embedding (v) (Section IV-B)}$
\ENDFOR
\STATE Calculate $L_{se}$ (Equation 31).
\ENDWHILE
\end{algorithmic}
\caption{\model{}}
\label{algo:Fuzzycone}
\end{algorithm}
\begin{algorithm}[!t]
\textbf{Input}: MMKG - $G(\mathcal{V},\mathcal{R},\mathcal{U}, \mathcal{M})$ \\
\textbf{Output}: Training Query and Answer Set\\
\begin{algorithmic}[1]
\FOR{each image $i$ of MMKG in the training set}
\STATE $\text{Generate scene graph } SG_i$
\ENDFOR
\STATE $\text{Merge MMKG, and $SG_i$'s by link prediction}$ 
$\text{using ComplEx}$
\FOR{each query structure (Figure 7)}
\STATE $\text{Generate $n$ queries via backtracking~\cite{ren2020beta}.} $ 
\ENDFOR
\end{algorithmic}
\caption{Query Generation}
\label{algo:query_gen}
\end{algorithm}
\section{Experiments}
\begin{figure*}[!t]
	\centering
	\includegraphics[width=0.8\paperwidth]{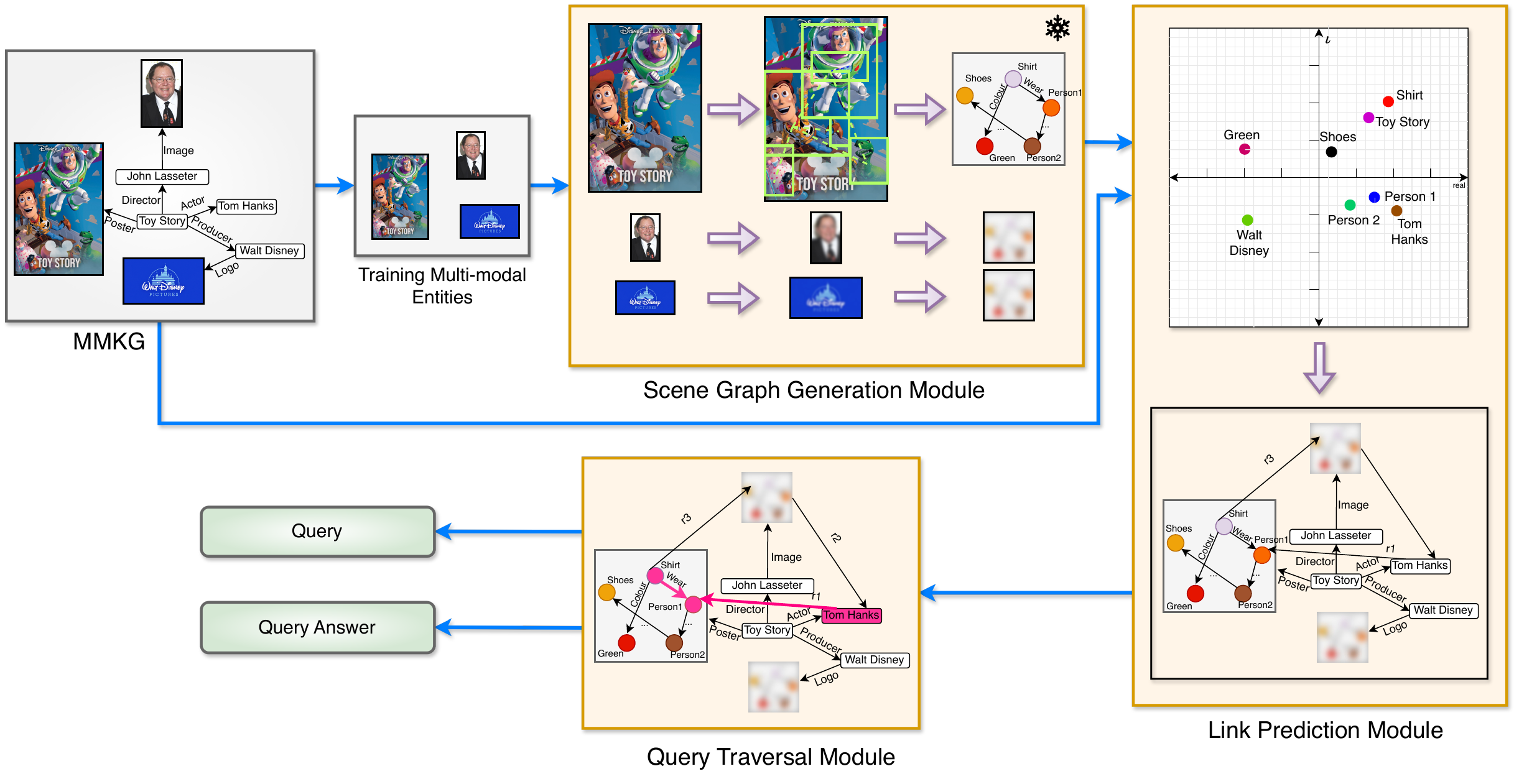}
	\caption{\textbf{Query Generation}: The Scene Graph Generation module takes all the image entities from the Multi-Modal Knowledge Graph (MMKG) as input and generates a scene graph for each entity. Link Prediction Module predicts links between these scene graphs and the original MMKG. Query Traversal Module generates queries for multiple query structures based on the entire knowledge graph (actual MMKG and scene graphs merged).}
	\label{fig:logical_query}
\end{figure*}
\begin{table*}[!ht]
    \caption{Average HITS@3 (\%) score for candidate multi-modal entities (first five rows of each dataset) and answer sub-entities for Type (I) answers for \model{} with and without graph transformer (grey rows). Bold results indicate the best score.}
    \centering
    \begin{tabular}{l|c|ccc|cc|cc|cc|ccc|cc|c}
    \toprule
       \textbf{Dataset} & \textbf{Model} & \textbf{1p} & \textbf{2p} & \textbf{3p} & \textbf{2i} & \textbf{3i} & \textbf{pi} & \textbf{ip} & \textbf{2u} & \textbf{up} & \textbf{pni} & \textbf{pin} & \textbf{inp} & \textbf{2in} & \textbf{3in} & \textbf{AVG} \\
         \midrule
        \multirow{7}{*}{FB15k} & BetaE & 5.6 & 1.7 & 0.4 & 7.0 & 7.8 & 5.7 & 1.7 & 5.8 & 1.7 & 4.3 & 0.5 & 0.3 & 4.3 & 4.6 & 3.7 \\
        & ConE & 8.8 & 4.1 & 2.3 & 11 & 8.7 & 7.7 & 4.3 & 8.2 & 4.8 & 5.4 & 2.7 & 1.7 & 4.6 & 5.3 & 5.7 \\
        & CylE & 20 & 8.9 & 7.5 & 24 & 27 & 19 & 11 & 23 & 9.5 & 17 & 6.6 & 3.6 & 15 & 20 & 15 \\
        & GNN-QE & 16 & 4.3 & 0.9 & 29 & 47 & 22 & 9.3 & 18 & 4.8 & 11 & 2.0 & 3.1 & 8.9 & 17 & 14 \\
        & \model{} (1a) & \textbf{93} & \textbf{45} & \textbf{30} & \textbf{96} & \textbf{97} & \textbf{74} & \textbf{36} & \textbf{87} & \textbf{44} & \textbf{83} & \textbf{39} & \textbf{20} & \textbf{88} & \textbf{80} & \textbf{67} \\
        & \mycc \model{} (1b) & \mycc 83 & \mycc 79 & \mycc 67 & \mycc 77 & \mycc 76 & \mycc 77 & \mycc 84 & \mycc 87 & \mycc 79 & \mycc 66 & \mycc 69 & \mycc 82 & \mycc 76 & \mycc 61 & \mycc 76 \\
        & \mycc \model{} (1b) w/o trans & \mycc 56 & \mycc 60 & \mycc 51 & \mycc 54 & \mycc 61 & \mycc 61 & \mycc 67 & \mycc 64 & \mycc 61 & \mycc 55 & \mycc 54 & \mycc 62 & \mycc 63 & \mycc 53 & \mycc 59 \\
        \midrule
        \multirow{7}{*}{FB15k-(237)} & BetaE &  34 & 23 & 11 & 36 & 37 & 33 & 16 & 30 & 19 & 27 & 7.8 & 6.5 & 21 & 24 & 23 \\
       & ConE & 10 & 11 & 8.0 & 8.4 & 9.4 & 12 & 11 & 10 & 11 & 6.5 & 5.5 & 3.6 & 4.7 & 6.1 & 8.0 \\
        & CylE & 57 & 45 & 31 & 53 & 62 & 57 & 34 & 55 & 36 & 49 & 18 & 14 & 40 & 47 & 43 \\
        & GNN-QE & 11 & 2.4 & 0.3 & 20 & 20 & 8.2 & 2.0 & 11 & 1.5 & 7.3 & 0.8 & 0.9 & 4.6 & 8.7 & 7.1 \\
       & \model{} (1a) & \textbf{99} & \textbf{72} & \textbf{48} & \textbf{99} & \textbf{99} & \textbf{96} & \textbf{58} & \textbf{99} & \textbf{66} & \textbf{97} & \textbf{53} & \textbf{27} & \textbf{98} & \textbf{97} & \textbf{80} \\
         & \mycc \model{} (1b) & \mycc 88 & \mycc 73 & \mycc 71 & \mycc 83 & \mycc 83 & \mycc 80 & \mycc 81 & \mycc 92 & \mycc 75 & \mycc 58 & \mycc 68 & \mycc 76 & \mycc 68 & \mycc 58 & \mycc 75 \\
         & \mycc \model{} (1b) w/o trans & \mycc 66 & \mycc 59 & \mycc 57 & \mycc 63 & \mycc 68 & \mycc 64 & \mycc 68 & \mycc 73 & \mycc 59 & \mycc 52 & \mycc 54 & \mycc 61 & \mycc 54 & \mycc 50 &
         \mycc 60 \\
         \midrule
        \multirow{7}{*}{YAGO15k} & BetaE & 5.6 & 0.5 & 0.8 & 4.5 & 1.7 & 1.7 & 1.0 & 1.9 & 0.8 & 1.6 & 0.5 & 0.7 & 1.6 & 0.6 & 1.7 \\
        & ConE & 1.3 & 0.2 & 0.7 & 0.2 & 0.3 & 0.2 & 0.2 & 0.8 & 0.7 & 0.1 & 0.1 & 0.4  & 0.2 & 0.0 & 0.4 \\
        & CylE & 7.9 & 1.5 & 2.2 & 5.4 & 3.1 & 3.0 & 2.1 & 5.2 & 1.5 & 4.3 & 1.6 & 1.4  & 3.9 & 0.9 & 3.1 \\
        & GNN-QE & 9.7 & 0.5 & 0.4 & 14 & 6.5 & 4.4 & 2.6 & 8.0 & 0.7 & 2.7 & 0.6 & 0.8 & 2.5 & 2.2 & 3.9 \\
       & \model{} (1a) & \textbf{89} & \textbf{43} & \textbf{25} & \textbf{96} & \textbf{79} & \textbf{55} & \textbf{34} & \textbf{70} & \textbf{32} & \textbf{73} & \textbf{25} & \textbf{16} & \textbf{78} & \textbf{54} & \textbf{59} \\
       & \mycc \model{} (1b) & \mycc 59 & \mycc 56 & \mycc 58 & \mycc 67 & \mycc 61 & \mycc 53 & \mycc 57 & \mycc 60 & \mycc 55 & \mycc 54 & \mycc 51 & \mycc 53 & \mycc 59 & \mycc 55 & \mycc 57 \\
       & \mycc \model{} (1b) w/o trans & \mycc 41 & \mycc 40 & \mycc 42 & \mycc 45 & \mycc 43 & \mycc 36 & \mycc 46 & \mycc 42 & \mycc 42 & \mycc 37 & \mycc 38 & \mycc 39 & \mycc 42 & \mycc 33 & \mycc 40 \\
       \midrule
        \multirow{7}{*}{DB15k} & BetaE & 24 & 14 & 7.6 & 27 & 27 & 21 & 14 & 18 & 13 & 20 & 12 & 10 & 19 & 19 & 18 \\
        & ConE & 9.4 & 12 & 8.4 & 8.7 & 9.1 & 11 & 12 & 8.5 & 10 & 5.6 & 7.9 & 7.7 & 6.0 & 7.1 & 8.9 \\
        & CylE & 44 & 27 & 20 & 46 & 48 & 33 & 24 & 38 & 21 & 39 & 20 & 17 & 41 & 42 & 33 \\
        & GNN-QE & 26 & 6.9 & 1.0 & 48 & 58 & 27 & 9.4 & 30 & 3.8 & 16 & 2.1 & 2.7 & 11 & 27 & 19 \\
       & \model{} (1a) & \textbf{86} & \textbf{39} & \textbf{22} & \textbf{96} & \textbf{98} & \textbf{81} & \textbf{37} & \textbf{82} & \textbf{37} & \textbf{83} & \textbf{34} & \textbf{21} & \textbf{84} & \textbf{93} & \textbf{69} \\
       & \mycc \model{} (1b) & \mycc 53 & \mycc 49 & \mycc 51 & \mycc 52 & \mycc 53 & \mycc 47 & \mycc 49 & \mycc 59 & \mycc 50 & \mycc 48 & \mycc 46 & \mycc 47 & \mycc 47 & \mycc 49 & \mycc 50 \\
       & \mycc \model{} (1b) w/o trans & \mycc 41 & \mycc 38 & \mycc 41 & \mycc 47 & \mycc 46 & \mycc 40 & \mycc 42 & \mycc 48 & \mycc 41 & \mycc 40 & \mycc 35 & \mycc 37 & \mycc 39 & \mycc 40 & \mycc 41 \\
       \bottomrule
    \end{tabular}
    \label{tab:mm_hits3_multi-modal}
\end{table*}
\begin{table*}[!ht]
    \caption{Average HITS@3 (\%) score for non-multi-modal query answers (Type (II) answers). Bold results indicate the best score.}
    \centering
    \begin{tabular}{l|c|ccc|cc|cc|cc|ccc|cc|c}
    \toprule
       \textbf{Dataset} & \textbf{Model} & \textbf{1p} & \textbf{2p} & \textbf{3p} & \textbf{2i} & \textbf{3i} & \textbf{pi} & \textbf{ip} & \textbf{2u} & \textbf{up} & \textbf{pni} & \textbf{pin} & \textbf{inp} & \textbf{2in} & \textbf{3in} & \textbf{AVG} \\
         \midrule
        \multirow{5}{*}{FB15k} & BetaE & 47 & 13 & 12 & 38 & 53 & 28 & 13 & 16 & 12 & 6.4 & 3.7 & 6.7 & 7.7 & 8.6 & 19 \\
        & ConE & 77 & 29 & 24 & 63 & 75 & 49 & 32 & 46 & 27 & 14 & 7.4 & 11 & 16 & 16 & 35 \\
        & CylE & 83 & 33 & 28 & 66 & 78 & 53 & 39 & 53 & 30 & 13 & 6.5 & 12 & 15 & 15 & 37 \\
        & GNN-QE & \textbf{85} & \textbf{60} & \textbf{46} & \textbf{80} & \textbf{86} & \textbf{72} & \textbf{61} & \textbf{70} & \textbf{51} & \textbf{37} & \textbf{34} & \textbf{38} & \textbf{48} & \textbf{46} & \textbf{58} \\
        & \model{} (2) & 64 & 25 & 23 & 52 & 65 & 41 & 28 & 28 & 23 & 10 & 5.8 & 10 & 11 & 11 & 28 \\
        \midrule
        \multirow{5}{*}{FB15k-(237)} & BetaE & 41 & 10 & 9.0 & 28 & 42 & 22 & 12 & 11 & 9.6 & 2.7 & 2.4 & 6.5 & 4.0 & 6.2 & 15 \\ 
        & ConE & 46 & 12 & 10 & 34 & 49 & 26 & 15 & 13 & 11 & 3.0 & 3.5 & 7.5 & 4.6 & 8.3 & 17 \\
        & CylE & \textbf{47} & \textbf{13} & \textbf{11} & 37 & 50 & 29 & 17 & \textbf{14} & \textbf{12} & 2.4 & 2.6 & 7.7 & 3.7 & 7.4 & 18 \\
        & GNN-QE & 45 & \textbf{13} & \textbf{11} & \textbf{42} & \textbf{58} & \textbf{33} & \textbf{20} & \textbf{14} & \textbf{12} & \textbf{5.1} & \textbf{6.1} & \textbf{8.8} & \textbf{6.1} & \textbf{16} & \textbf{21} \\
        & \model{} (2)  & 43 & 11 & 10 & 32 & 45 & 23 & 14 & 12 & 10 & 3.2 & 2.8 & 7.2 & 5.2 & 8.1 & 16 \\
         \midrule
        \multirow{5}{*}{YAGO15k} & BetaE  & 84 & 66 & 48 & 94 & 97 & 75 & 59 & 87 & 63 & 63 & 50 & 37 & 77 & 75 & 70 \\
        & ConE & 92 & 82 & 71 & 98 & 99 & 82 & 65 & 97 & 81 & 74 & 70 & 52 & 90 & 87 & 81 \\
        & CylE & 94 & 87 & 78 & 99 & 99 & 85 & 72 & \textbf{99} & 85 & 71 & 71 & 55 & 86 & 84 & 83 \\
        & GNN-QE & \textbf{100} & \textbf{94} & \textbf{82} & \textbf{100} & \textbf{100} & \textbf{99} & \textbf{98} & \textbf{99} & \textbf{97} & \textbf{99} & \textbf{94} & \textbf{94} & \textbf{99} & \textbf{99} & \textbf{97} \\
        & \model{} (2) & 80 & 75 & 66 & 94 & 97 & 74 & 58 & 79 & 72 & 65 & 56 & 43 & 77 & 73 & 72 \\
         \midrule
        \multirow{5}{*}{DB15k} & BetaE & 83 & 59 & 36 & 96 & 98 & 85 & 67 & 92 & 60 & 80 & 45 & 37 & 83 & 75 & 71 \\
        & ConE & 89 & 70 & 50 & 98 & \textbf{99} & 91 & 73 & 98 & 76 & 87 & 62 & 47 & 92 & 86 & 80 \\
        & CylE & 91 & 77 & 58 & \textbf{99} & \textbf{99} & 92 & 81 & \textbf{99} & 82 & 86 & 64 & 55 & 87 & 83 & 82 \\
        & GNN-QE & \textbf{99} & \textbf{84} & \textbf{70} & \textbf{99} & \textbf{99} & \textbf{98} & \textbf{91} & \textbf{99} & \textbf{85} & \textbf{99} & \textbf{90} & \textbf{88} & \textbf{99} & \textbf{99} & \textbf{93} \\
        & \model{} (2) & 80 & 62 & 44 & 95 & 98 & 83 & 64 & 84 & 64 & 76 & 46 & 36 & 80 & 67 & 70 \\
       \bottomrule
    \end{tabular}
    \label{tab:nmm_hits3_multi-modal}
\end{table*}
\begin{table*}[!ht]
    \caption{MRR (\%) score FB15k (NMM) dataset. Bold results indicate the best score.}
    \centering
    \begin{tabular}{l|ccc|cc|cc|cc|ccc|cc|c}
    \toprule
         \textbf{Model} & \textbf{1p} & \textbf{2p} & \textbf{3p} & \textbf{2i} & \textbf{3i} & \textbf{pi} & \textbf{ip} & \textbf{2u} & \textbf{up} & \textbf{pni} & \textbf{pin} & \textbf{inp} & \textbf{2in} & \textbf{3in} & \textbf{AVG} \\
         \midrule
          BetaE & 64 & 24 & 24 & 55 & 66 & 43 & 27 & 40 & 24 & 12 & 6.1 & 11 & 13 & 14 & 30 \\
          ConE & 73 & 33 & 28 & 64 & 73 & 50 & 35 & 54 & 31 & 14 & 9.5 & 12 & 17 & 18 & 37 \\
          \model & 73 & 33 & 29 & 64 & 72 & 50 & 34 & 53 & 31 & 14 & 9.0 & 12 & 17 & 18 & 36 \\
          CylE & 78 & 37 & 30 & 67 & 75 & 53 & 40 & 59 & 33 & 14 & 7.8 & 13 & 15 & 16 & 38 \\
          GNN-QE & \textbf{87} & \textbf{69} & \textbf{59} & \textbf{81} & \textbf{85} & \textbf{71} & \textbf{68} & \textbf{75} & \textbf{60} & \textbf{34} & \textbf{30} & \textbf{42} & \textbf{47} & \textbf{45} & \textbf{61} \\
          \bottomrule
    \end{tabular}
    \label{tab:mrr_nmulti-modal}
\end{table*}

\end{document}